# Learning to Win by Reading Manuals
# in a Monte-Carlo Framework

**S.R.K. Branavan**                                    BRANAVAN@CSAIL.MIT.EDU
*Computer Science and Artificial Intelligence Laboratory*
*Massachusetts Institute of Technology*

**David Silver**                                            D.SILVER@CS.UCL.AC.UK
*Department of Computer Science*
*University College London*

**Regina Barzilay**                                        REGINA@CSAIL.MIT.EDU
*Computer Science and Artificial Intelligence Laboratory*
*Massachusetts Institute of Technology*

## Abstract

Domain knowledge is crucial for effective performance in autonomous control systems. Typically, human effort is required to encode this knowledge into a control algorithm. In this paper, we present an approach to language grounding which automatically interprets text in the context of a complex control application, such as a game, and uses domain knowledge extracted from the text to improve control performance. Both text analysis and control strategies are learned jointly using only a feedback signal inherent to the application. To effectively leverage textual information, our method automatically extracts the text segment most relevant to the current game state, and labels it with a task-centric predicate structure. This labeled text is then used to bias an action selection policy for the game, guiding it towards promising regions of the action space. We encode our model for text analysis and game playing in a multi-layer neural network, representing linguistic decisions via latent variables in the hidden layers, and game action quality via the output layer. Operating within the Monte-Carlo Search framework, we estimate model parameters using feedback from simulated games. We apply our approach to the complex strategy game Civilization II using the official game manual as the text guide. Our results show that a linguistically-informed game-playing agent significantly outperforms its language-unaware counterpart, yielding a 34% absolute improvement and winning over 65% of games when playing against the built-in AI of Civilization.

## 1. Introduction

In this paper, we study the task of grounding document content in control applications such as computer games. In these applications, an agent attempts to optimize a utility function (e.g., game score) by learning to select situation-appropriate actions. In complex domains, finding a winning strategy is challenging even for humans. Therefore, human players typically rely on manuals and guides that describe promising tactics and provide general advice about the underlying task. Surprisingly, such textual information has never been utilized in control algorithms despite its potential to greatly improve performance. Our goal, therefore, is to develop methods that can achieve this in an automatic fashion.

> The natural resources available where a population settles affects its ability to produce food and goods. Cities built on or near water sources can irrigate to increase their crop yields, and cities near mineral resources can mine for raw materials. Build your city on a plains or grassland square with a river running through it if possible.

Figure 1: An excerpt from the user manual of the game Civilization II.

We explore this question in the context of strategy games, a challenging class of large scale adversarial planning problems.

Consider for instance the text shown in Figure 1. This is an excerpt from the user manual of the game Civilization II.[1] This text describes game locations where the action *build-city* can be effectively applied. A stochastic player that does not have access to this text would have to gain this knowledge the hard way: it would repeatedly attempt this action in a myriad of states, thereby learning the characterization of promising state-action pairs based on observed game outcomes. In games with large state spaces, long planning horizons, and high-branching factors, this approach can be prohibitively slow and ineffective. An algorithm with access to the text, however, could learn correlations between words in the text and game attributes – e.g., the word "river" and places with rivers in the game – thus leveraging strategies described in text to select better actions.

To improve the performance of control applications using domain knowledge automatically extracted from text, we need to address the following challenges:

- **Grounding Text in the State-Action Space of a Control Application** Text guides provide a wealth of information about effective control strategies, including situation-specific advice as well as general background knowledge. To benefit from this information, an algorithm has to learn the mapping between the text of the guide, and the states and actions of the control application. This mapping allows the algorithm to find state-specific advice by matching state attributes to their verbal descriptions. Furthermore, once a relevant sentence is found, the mapping biases the algorithm to select the action proposed in the guide document. While this mapping can be modeled at the word-level, ideally we would also use information encoded in the structure of the sentence – such as the predicate argument structure. For instance, the algorithm can explicitly identify predicates and state attribute descriptions, and map them directly to structures inherent in the control application.

- **Annotation-free Parameter Estimation** While the above text analysis tasks relate to well-known methods in information extraction, prior work has primarily focused on supervised methods. In our setup, text analysis is state dependent, therefore annotations need to be representative of the entire state space. Given an enormous state space that continually changes as the game progresses, collecting such annotations is impractical. Instead, we propose to learn text analysis based on a feedback signal inherent to the control application, e.g., the game score. This feedback is computed automatically at each step of the game, thereby allowing the algorithm to continuously adapt to the local, observed game context.

---

1. http://en.wikipedia.org/wiki/Civilization_II

- **Effective Integration of Extracted Text Information into the Control Application** Most text guides do not provide complete, step-by-step advice for all situations that a player may encounter. Even when such advice is available, the learned mapping may be noisy, resulting in suboptimal choices. Therefore, we need to design a method which can achieve effective control in the absence of textual advice, while robustly integrating automatically extracted information when available. We address this challenge by incorporating language analysis into *Monte-Carlo Search*, a state-of-the-art framework for playing complex games. Traditionally this framework operates only over state and action features. By extending Monte-Carlo search to include textual features, we integrate these two sources of information in a principled fashion.

## 1.1 Summary of the Approach

We address the above challenges in a unified framework based on *Markov Decision Processes* (MDP), a formulation commonly used for game playing algorithms. This setup consists of a game in a stochastic environment, where the goal of the player is to maximize a given utility function $R(s)$ at state $s$. The player's behavior is determined by an action-value function $Q(s, a)$ that assesses the goodness of action $a$ at state $s$ based on the attributes of $s$ and $a$.

To incorporate linguistic information into the MDP formulation, we expand the action value function to include linguistic features. While state and action features are known at each point of computation, relevant words and their semantic roles are not observed. Therefore, we model text relevance as a hidden variable. Similarly, we use hidden variables to discriminate the words that describe *actions* and those that describe *state attributes* from the rest of the sentence. To incorporate these hidden variables in our action-value function, we model $Q(s, a)$ as a non-linear function approximation using a multi-layer neural network.

Despite the added complexity, all the parameters of our non-linear model can be effectively learned in the Monte-Carlo Search framework. In Monte-Carlo Search, the action-value function is estimated by playing multiple simulated games starting at the current game state. We use the observed reward from these simulations to update the parameters of our neural network via backpropagation. This focuses learning on the current game state, allowing our method to learn language analysis and game-play appropriate to the observed game context.

## 1.2 Evaluation

We test our method on the strategy game Civilization II, a notoriously challenging game with an immense action space.[2] As a source of knowledge for guiding our model, we use the official game manual. As a baseline, we employ a similar Monte-Carlo search based player which does not have access to textual information. We demonstrate that the linguistically-informed player significantly outperforms the baseline in terms of the number of games won. Moreover, we show that modeling the deeper linguistic structure of sentences further improves performance. In full-length games, our algorithm yields a 34% improve-ment over a language unaware baseline and wins over 65% of games against the built-in, hand-crafted AI of Civilization II. A video of our method playing the game is available at http://groups.csail.mit.edu/rbg/code/civ/video. The code and data for this work, along with a complete experimental setup and a preconfigured environment in a virtual machine are available at http://groups.csail.mit.edu/rbg/code/civ.

---

2. Civilization II was #3 in IGN's 2007 list of top video games of all time.
(http://top100.ign.com/2007/ign_top_game_3.html)

### 1.3 Roadmap

In Section 2, we provide intuition about the benefits of integrating textual information into learning algorithms for control. Section 3 describes prior work on language grounding, emphasizing the unique challenges and opportunities of our setup. This section also positions our work in a large body of research on Monte-Carlo based players. Section 4 presents background on Monte-Carlo Search as applied to game playing. In Section 5 we present a multi-layer neural network formulation for the action-value function that combines information from the text and the control application. Next, we present a Monte-Carlo method for estimating the parameters of this non-linear function. Sections 6 and 7 focus on the application of our algorithm to the game Civilization II. In Section 8 we compare our method against a range of competitive game-playing baselines, and empirically analyze the properties of the algorithm. Finally, in Section 9 we discuss the implications of this research, and conclude.

## 2. Learning Game Play from Text

In this section, we provide an intuitive explanation of how textual information can help improve action selection in a complex game. For clarity, we first discuss the benefits of textual information in the supervised scenario, thereby decoupling questions concerning modeling and representation from those related to parameter estimation. Assume that every state $s$ is represented by a set of $n$ features $[s_1, s_2, \ldots, s_n]$. Given a state $s$, our goal is to select the best possible action $a_j$ from a fixed set $A$. We can model this task as multiclass classification, where each choice $a_j$ is represented by a feature vector $[(s_1, a_j), (s_2, a_j), \ldots, (s_n, a_j)]$. Here, $(s_i, a_j), i \in [1, n]$ represents a feature created by taking the Cartesian product between $[s_1, s_2, \ldots, s_n]$ and $a_j$. To learn this classifier effectively, we need a training set that sufficiently covers the possible combinations of state features and actions. However, in domains with complex state spaces and a large number of possible actions, many instances of state-action feature values will be unobserved in training.

Now we show how the generalization power of the classifier can be improved using textual information. Assume that each training example, in addition to a state-action pair, contains a sentence that may describe the action to be taken given the state attributes. Intuitively, we want to enrich our basic classifier with features that capture the correspondence between states and actions, and words that describe them. Given a sentence $w$ composed of word types $w_1, w_2, \ldots, w_m$, these features can be of the form $(s_i, w_k)$ and $(a_j, w_k)$ for every $i \in [1, n]$, $k \in [1, m]$ and $a_j \in A$. Assuming that an action is described using similar words throughout the guide, we expect that a text-enriched classifier would be able to learn this correspondence via the features $(a_j, w_k)$. A similar intuition holds for learning the correspondence between state-attributes and their descriptions represented by features $(s_i, w_k)$. Through these features, the classifier can connect state $s$ and action $a_j$ based on the evidence provided in the guiding sentence and their occurrences in other contexts throughout the training data. A text-free classifier may not support such an association if the action does not appear in a similar state context in a training set.

The benefits of textual information extend to models that are trained using control feedback rather than supervised data. In this training scenario, the algorithm assesses the goodness of a given state-action combination by simulating a limited number of game turns after the action is taken and observing the control feedback provided by the underlying application. The algorithm has a built-in mechanism (see Section 4) that employs the observed feedback to learn feature weights, and intelligently samples the space in search for promising state-action pairs. When the algorithm has access to a collection of sentences, a similar feedback-based mechanism can be used to find sentences that match a given state-action pair (Section 5.1). Through the state- and action-description features ($s_i, w_k$) and ($a_j, w_k$), the algorithm jointly learns to identify relevant sentences and to map actions and states to their descriptions. Note that while we have used classification as the basis of discussion in this section, in reality our methods will learn a regression function.

## 3. Related Work

In this section, we first discuss prior work in the field of grounded language acquisition. Subsequently we look are two areas specific to our application domain – i.e., natural language analysis in the context of games, and Monte-Carlo Search applied to game playing.

### 3.1 Grounded Language Acquisition

Our work fits into the broad area of research on grounded language acquisition where the goal is to learn linguistic analysis from a non-linguistic situated context (Oates, 2001; Barnard & Forsyth, 2001; Siskind, 2001; Roy & Pentland, 2002; Yu & Ballard, 2004; Chen & Mooney, 2008; Zettlemoyer & Collins, 2009; Liang, Jordan, & Klein, 2009; Branavan, Chen, Zettlemoyer, & Barzilay, 2009; Branavan, Zettlemoyer, & Barzilay, 2010; Vogel & Jurafsky, 2010; Clarke, Goldwasser, Chang, & Roth, 2010; Tellex, Kollar, Dickerson, Walter, Banerjee, Teller, & Roy, 2011; Chen & Mooney, 2011; Liang, Jordan, & Klein, 2011; Goldwasser, Reichart, Clarke, & Roth, 2011). The appeal of this formulation lies in reducing the need for manual annotations, as the non-linguistic signals can provide a powerful, albeit noisy, source of supervision for learning. In a traditional grounding setup it is assumed that the non-linguistic signals are parallel in content to the input text, motivating a machine translation view of the grounding task. An alternative approach models grounding in the control framework where the learner actively acquires feedback from the non-linguistic environment and uses it to drive language interpretation. Below we summarize both approaches, emphasizing the similarity and differences with our work.

#### 3.1.1 LEARNING GROUNDING FROM PARALLEL DATA

In many applications, linguistic content is tightly linked to perceptual observations, providing a rich source of information for learning language grounding. Examples of such parallel data include images with captions (Barnard & Forsyth, 2001), Robocup game events paired with a text commentary (Chen & Mooney, 2008), and sequences of robot motor actions de-scribed in natural language (Tellex et al., 2011). The large diversity in the properties of such parallel data has resulted in the development of algorithms tailored for specific grounding contexts, instead of an application-independent grounding approach. Nevertheless, existing grounding approaches can be characterized along several dimensions that illuminate the connection between these algorithms:

- **Representation of Non-Linguistic Input** The first step in grounding words in perceptual data is to discretize the non-linguistic signal (e.g., an image) into a representation that facilitates alignment. For instance, Barnard and Forsyth (2001) segment images into regions that are subsequently mapped to words. Other approaches intertwine alignment and segmentation into a single step (Roy & Pentland, 2002), as the two tasks are clearly interrelated. In our application, segmentation is not required as the state-action representation is by nature discrete.

  Many approaches move beyond discretization, aiming to induce rich hierarchical structures over the non-linguistic input (Fleischman & Roy, 2005; Chen & Mooney, 2008, 2011). For instance, Fleischman and Roy (2005) parse action sequences using a context-free grammar which is subsequently mapped into semantic frames. Chen and Mooney (2008) represent action sequences using first order logic. In contrast, our algorithm capitalizes on the structure readily available in our data – state-action transitions. While inducing a richer structure on the state-action space may benefit mapping, it is a difficult problem in its own right from the field of hierarchical planning (Barto & Mahadevan, 2003).

- **Representation of Linguistic Input** Early grounding approaches used the bag-of-words approach to represent input documents (Yu & Ballard, 2004; Barnard & Forsyth, 2001; Fleischman & Roy, 2005). More recent methods have relied on a richer representation of linguistic data, such as syntactic trees (Chen & Mooney, 2008) and semantic templates (Tellex et al., 2011). Our method incorporates linguistic information at multiple levels, using a feature-based representation that encodes both words as well as syntactic information extracted from dependency trees. As shown by our results, richer linguistic representations can significantly improve model performance.

- **Alignment** Another common feature of existing grounding models is that the training procedure crucially depends on how well words are aligned to non-linguistic structures. For this reason, some models assume that alignment is provided as part of the training data (Fleischman & Roy, 2005; Tellex et al., 2011). In other grounding algorithms, the alignment is induced as part of the training procedure. Examples of such approaches are the methods of Barnard and Forsyth (2001), and Liang et al. (2009). Both of these models jointly generate the text and attributes of the grounding context, treating alignment as an unobserved variable.

  In contrast, we do not explicitly model alignment in our model due to the lack of parallel data. Instead, we aim to extract relevant information from text and infuse it into a control application.

### 3.1.2 Learning Grounding from Control Feedback

More recent work has moved away from the reliance on parallel corpora, using control feedback as the primary source of supervision. The assumption behind this setup is that when textual information is used to drive a control application, the application's performance will correlate with the quality of language analysis. It is also assumed that the performance measurement can be obtained automatically. This setup is conducive to reinforcement learning approaches which can estimate model parameters from the feedback signal, even it is noisy and delayed.

One line of prior work has focused on the task of mapping textual instructions into a *policy* for the control application, assuming that text fully specifies all the actions to be executed in the environment. For example, in our previous work (Branavan et al., 2009, 2010), this approach was applied to the task of translating instructions from a computer manual to executable GUI actions. Vogel and Jurafsky (2010) demonstrate that this grounding framework can effectively map navigational directions to the corresponding path in a map. A second line of prior work has focused on full semantic parsing – converting a given text into a formal meaning representation such as first order logic (Clarke et al., 2010). These methods have been applied to domains where the correctness of the output can be accurately evaluated based on control feedback – for example, where the output is a database query which when executed provides a clean, oracle feedback signal for learning. This line of work also assumes that the text fully specifies the required output.

While our method is also driven by control feedback, our language interpretation task itself is fundamentally different. We assume that the given text document provides high-level advice without directly describing the correct actions for every potential game state. Furthermore, the textual advice does not necessarily translate to a single strategy – in fact, the text may describe several strategies, each contingent on specific game states. For this reason, the strategy text cannot simply be interpreted directly into a policy. Therefore, our goal is to bias a learned policy using information extracted from text. To this end, we do not aim to achieve a complete semantic interpretation, but rather use a partial text analysis to compute features relevant for the control application.

## 3.2 Language Analysis and Games

Even though games can provide a rich domain for situated text analysis, there have only been a few prior attempts at leveraging this opportunity (Gorniak & Roy, 2005; Eisenstein, Clarke, Goldwasser, & Roth, 2009).

Eisenstein et al. (2009) aim to automatically extract information from a collection of documents to help identify the rules of a game. This information, represented as predicate logic formulae, is estimated in an unsupervised fashion via a generative model. The extracted formulae, along with observed traces of game play are subsequently fed to an Inductive Logic Program, which attempts to reconstruct the rules of the game. While at the high-level, our goal is similar, i.e., to extract information from text useful for an external task, there are several key differences. Firstly, while Eisenstein et al. (2009) analyze the text and the game as two disjoint steps, we model both tasks in an integrated fashion. This allows our model to learn a text analysis pertinent to game play, while at the same time using text to guide game play. Secondly, our method learns both text analysis and game play from a feedback signal inherent to the game, avoiding the need for pre-compiled game traces. This enables our method to operate effectively in complex games where collecting a sufficiently representative set of game traces can be impractical.

Gorniak and Roy (2005) develop a machine controlled game character which responds to spoken natural language commands. Given traces of game actions manually annotated with transcribed speech, their method learns a structured representation of the text and aligned action sequences. This learned model is then used to interpret spoken instructions by grounding them in the actions of a human player and the current game state. While the method itself does not learn to play the game, it enables human control of an additional game character via speech. In contrast to Gorniak and Roy (2005), we aim to develop algorithms to fully and autonomously control all actions of one player in the game. Furthermore, our method operates on the game's user manual rather than on human provided, contextually relevant instructions. This requires our model to identify if the text contains information useful in the current game state, in addition to mapping the text to productive actions. Finally, our method learns from game feedback collected via active interaction without relying on manual annotations. This allows us to effectively operate on complex games where collecting traditional labeled traces would be prohibitively expensive.

## 3.3 Monte-Carlo Search for Game AI

Monte-Carlo Search (MCS) is a state-of-the-art framework that has been very successfully applied, in prior work, to playing complex games such as Go, Poker, Scrabble, and real-time strategy games (Gelly, Wang, Munos, & Teytaud, 2006; Tesauro & Galperin, 1996; Billings, Castillo, Schaeffer, & Szafron, 1999; Sheppard, 2002; Schäfer, 2008; Sturtevant, 2008; Balla & Fern, 2009). This framework operates by playing simulated games to estimate the goodness or *value* of different candidate actions. When the game's state and action spaces are complex, the number of simulations needed for effective play become prohibitively large. Previous application of MCS have addressed this issue using two orthogonal techniques: (1) they leverage domain knowledge to either guide or prune action selection, (2) they estimate the value of untried actions based on the observed outcomes of simulated games. This estimate is then used to bias action selection. Our MCS based algorithm for games relies on both of the above techniques. Below we describe the differences between our application of these techniques and prior work.

### 3.3.1 LEVERAGING DOMAIN KNOWLEDGE

Domain knowledge has been shown to be critically important to achieving good performance from MCS in complex games. In prior work this has been achieved by manually encoding relevant domain knowledge into the game playing algorithm – for example, via manually specified heuristics for action selection (Billings et al., 1999; Gelly et al., 2006), hand crafted features (Tesauro & Galperin, 1996), and value functions encoding expert knowledge (Sturtevant, 2008). In contrast to such approaches, our goal is to automatically extract and use domain knowledge from relevant natural language documents, thus bypassing the need for manual specification. Our method learns both text interpretation and game action selection based on the outcomes of simulated games in MCS. This allows it to identify and leverage textual domain knowledge relevant to the observed game context.

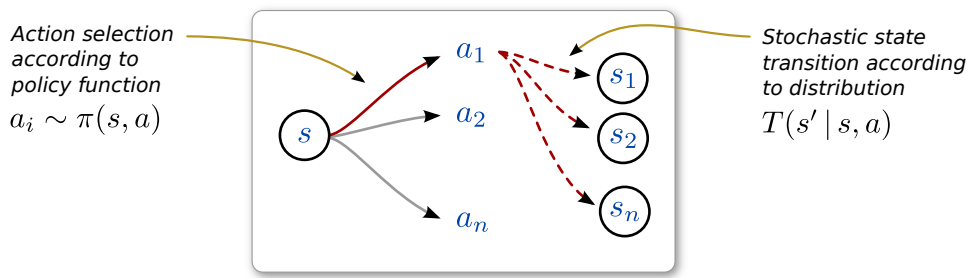

Figure 2: Markov Decision Process. Actions are selected according to *policy function* $\pi(s, a)$ given the current state $s$. The execution of the selected action $a_i$ (e.g., $a_1$), causes the MDP to transition to a new state $s'$ according to the stochastic state transition distribution $T(s' \mid s, a)$.

### 3.3.2 Estimating the Value of Untried Actions

Previous approaches to estimating the value of untried actions have relied on two techniques. The first, *Upper Confidence bounds for Tree* (UCT) is a heuristic used in concert with the Monte-Carlo Tree Search variant of MCS. It augments an action's value with an exploration bonus for rarely visited state-action pairs, resulting in better action selection and better overall game performance (Gelly et al., 2006; Sturtevant, 2008; Balla & Fern, 2009). The second technique is to learn a linear function approximation of action values for the current state $s$, based on game feedback (Tesauro & Galperin, 1996; Silver, Sutton, & Müller, 2008). Even though our method follows the latter approach, we model action-value $Q(s, a)$ via a non-linear function approximation. Given the complexity of our application domain, this non-linear approximation generalizes better than a linear one, and as shown by our results significantly improves performance. More importantly, the non-linear model enables our method to represent text analysis as latent variables, allowing it to use textual information to estimate the value of untried actions.

## 4. Monte-Carlo Search

Our task is to leverage textual information to help us win a turn-based strategy game against a given opponent. In this section, we first describe the *Monte-Carlo Search* framework within which our method operates. The details of our linguistically informed Monte-Carlo Search algorithm are given in Section 5.

### 4.1 Game Representation

Formally, we represent the given turn-based stochastic game as a Markov Decision Process (MDP). This MDP is defined by the 4-tuple $\langle S, A, T, R \rangle$, where

- *State space*, $S$, is the set of all possible states. Each state $s \in S$ represents a complete configuration of the game in-between player turns.

- *Action space*, $A$, is the set of all possible actions. In a turn-based strategy game, a player controls multiple game units at each turn. Thus, each action $a \in A$ represents the joint assignment of all unit actions executed by the current player during the turn.

- *Transition distribution*, $T(s' \mid s, a)$, is the probability that executing action $a$ in state $s$ will result in state $s'$ at the next game turn. This distribution encodes the way the game state changes due to both the game rules, and the opposing player's actions. For this reason, $T(s' \mid s, a)$ is stochastic – as shown in Figure 2, executing the same action $a$ at a given state $s$ can result in different outcomes $s'$.

- *Reward function*, $R(s) \in \mathbb{R}$, is the immediate reward received when transitioning to state $s$. The value of the reward correlates with the goodness of actions executed up to now, with higher reward indicating better actions.

All the above aspects of the MDP representation of the game – i.e., $S$, $A$, $T()$ and $R()$ – are defined implicitly by the game rules. At each step of the game, the game-playing agent can observe the current game state $s$, and has to select the best possible action $a$. When the agent executes action $a$, the game state changes according to the state transition distribution. While $T(s' \mid s, a)$ is not known a priori, state transitions can be sampled from this distribution by invoking the game code as a black-box simulator – i.e., by playing the game. After each action, the agent receives a reward according to the reward function $R(s)$. In a game playing setup, the value of this reward is an indication of the chances of winning the game from state $s$. Crucially, the reward signal may be delayed – i.e., $R(s)$ may have a non-zero value only for game ending states such as a win, a loss, or a tie.

The game playing agent selects actions according to a stochastic *policy* $\pi(s, a)$, which specifies the probability of selecting action $a$ in state $s$. The expected total reward after executing action $a$ in state $s$, and then following policy $\pi$ is termed the *action-value function* $Q^\pi(s, a)$. Our goal is to find the *optimal policy* $\pi^*(s, a)$ which maximizes the expected total reward – i.e., maximizes the chances of winning the game. If the *optimal action-value function* $Q^{\pi^*}(s, a)$ is known, the optimal game-playing behavior would be to select the action $a$ with the highest $Q^{\pi^*}(s, a)$. While it may be computationally hard to find an optimal policy $\pi^*(s, a)$ or $Q^{\pi^*}(s, a)$, many well studied algorithms are available for estimating an effective approximation (Sutton & Barto, 1998).

## 4.2 Monte-Carlo Framework for Computer Games

The Monte-Carlo Search algorithm, shown in Figure 3, is a simulation-based search paradigm for dynamically estimating the action-values $Q^\pi(s, a)$ for a given state $s_t$ (see Algorithm 1 for pseudo code). This estimate is based on the rewards observed during multiple *roll-outs*, each of which is a simulated game starting from state $s_t$.[3] Specifically, in each roll-out, the algorithm starts at state $s_t$, and repeatedly selects and executes actions according to a *simulation policy* $\pi(s, a)$, sampling state transitions from $T(s' \mid s, a)$. On game completion at time $\tau$, the final reward $R(s_\tau)$ is measured, and the action-value function is updated accordingly.[4] As in Monte-Carlo control (Sutton & Barto, 1998), the updated action-value function $Q^\pi(s, a)$ is used to define an improved simulation policy, thereby directing subsequent roll-outs towards higher scoring regions of the game state space. After a fixed number of roll-outs have been performed, the action with the highest average final reward in the simulations is selected and played in the actual game state $s_t$. This process is repeated for each state encountered during the actual game, with the action-value function being relearned from scratch for each new game state.[5] The simulation policy usually selects actions to maximize the action-value function. However, sometimes other valid actions are also randomly explored in case they are more valuable than predicted by the current es-timate of $Q^\pi(s, a)$. As the accuracy of $Q^\pi(s, a)$ improves, the quality of action selection improves and vice versa, in a cycle of continual improvement (Sutton & Barto, 1998).

---

3. Monte-Carlo Search assumes that it is possible to play simulated games. These simulations may be played against a heuristic AI player. In our experiments, the built-in AI of the game is used as the opponent.

4. In general, roll-outs are run until game completion. If simulations are expensive, as is the case in our domain, roll-outs can be truncated after a fixed number of steps. This however depends on the availability of an approximate reward signal at the truncation point. In our experiments, we use the built-in score of the game as the reward. This reward is noisy, but available at every stage of the game.

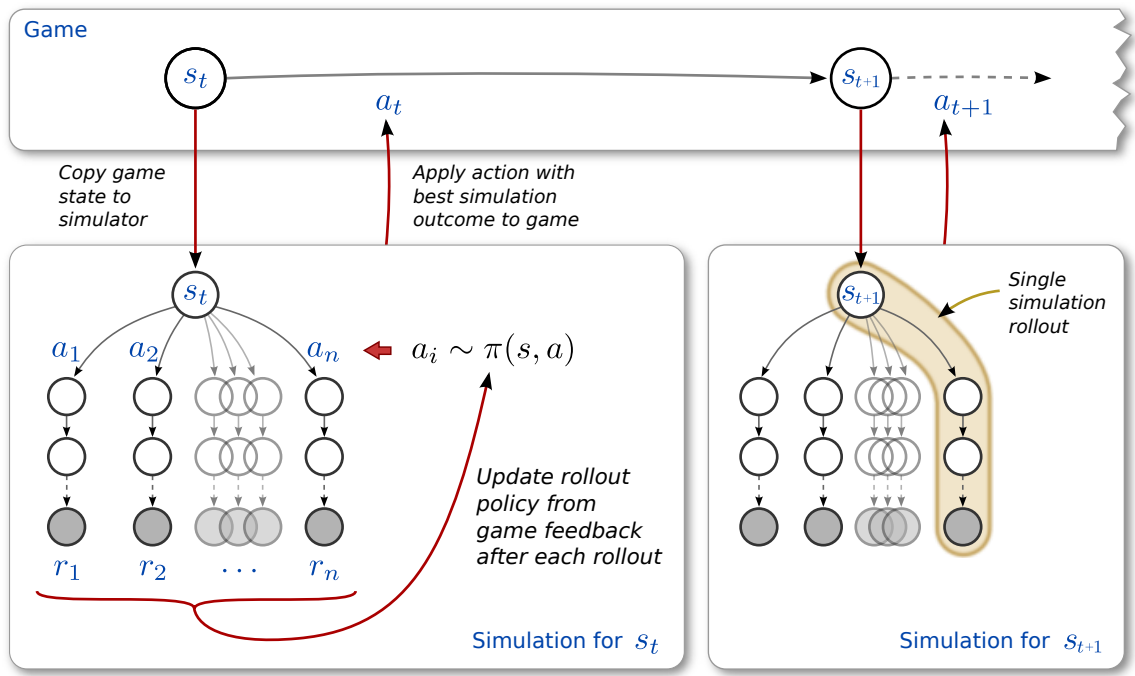

Figure 3: Overview of Monte-Carlo Search algorithm. For each game state $s_t$, an independent set of simulated games or *roll-outs* are done to find the best possible game action $a_t$. Each roll-out starts at state $s_t$, with actions selected according to a *simulation policy* $\pi(s, a)$. This policy is learned from the roll-outs themselves – with the roll-outs improving the policy, which in turn improves roll-out action selection. The process is repeated for every actual game state, with the simulation policy being relearned from scratch each time.

---

5. While it is conceivable that sharing the action-value function across the roll-outs of different game states would be beneficial, this was empirically not the case in our experiments. One possible reason is that in our domain, the game dynamics change radically at many points during the game – e.g., when a new technology becomes available. When such a change occurs, it may actually be detrimental to play according to the action-value function from the previous game step. Note however, that the action-value function is indeed shared across the roll-outs for a single game state $s_t$, with parameters updated by successive roll-outs. This is how the learned model helps improve roll-out action selection, and thereby improves game play. The setup of relearning from scratch for each game state has been shown to be beneficial even in stationary environments (Sutton, Koop, & Silver, 2007).

The success of Monte-Carlo Search depends on its ability to make a fast, local estimate of the action-value function from roll-outs collected via simulated play. However in games with large branching factors, it may not be feasible to collect sufficient roll-outs, especially when game simulation is computationally expensive. Thus it is crucial that the learned action-value function generalizes well from a small number of roll-outs – i.e., observed states, actions and rewards. One way to achieve this is to model the action-value function as a linear combination of state and action attributes:

$$Q^\pi(s, a) = \vec{w} \cdot \vec{f}(s, a).$$

Here $\vec{f}(s, a) \in \mathbb{R}^n$ is a real-valued feature function, and $\vec{w}$ is a weight vector. Prior work has shown such linear value function approximations to be effective in the Monte-Carlo Search framework (Silver et al., 2008).

Note that learning the action-value function $Q(s, a)$ in Monte-Carlo Search is related to Reinforcement Learning (RL) (Sutton & Barto, 1998). In fact, in our approach, we use standard gradient descent updates from RL to estimate the parameters of $Q(s, a)$. There is, however, one crucial difference between these two techniques: In general, the goal in RL is to find a $Q(s, a)$ applicable to *any* state the agent may observe during its existence. In the Monte-Carlo Search framework, the aim is to learn a $Q(s, a)$ specialized to the *current* state $s$. In essence, $Q(s, a)$ is relearned for every observed state in the actual game, using the states, actions and feedback from simulations. While such relearning may seem suboptimal, it has two distinct advantages: first, since $Q(s, a)$ only needs to model the current state, it can be representationally much simpler than a global action-value function. Second, due to this simpler representation, it can be learned from fewer observations than a global action-value function (Sutton et al., 2007). Both of these properties are important when the state space is extremely large, as is the case in our domain.

## 5. Adding Linguistic Knowledge to the Monte-Carlo Framework

The goal of our work is to improve the performance of the Monte-Carlo Search framework described above, using information automatically extracted from text. In this section, we describe how we achieve this in terms of model structure and parameter estimation.

### 5.1 Model Structure

To achieve our aim of leveraging textual information to improve game-play, our method needs to perform three tasks: (1) identify sentences relevant to the current game state, (2) label sentences with a predicate structure, and (3) predict good game actions by combining game features with text features extracted via the language analysis steps. We first describe how each of these tasks can be modeled separately before showing how we integrate them into a single coherent model.

---

**procedure**  PlayGame ()

---

*Initialize game state to fixed starting state*
$s_1 \leftarrow s_0$

**for** $t = 1 \ldots T$ **do**

> *Run N simulated games*
> **for** $i = 1 \ldots N$ **do**
> > $(a_i, r_i) \leftarrow$ SimulateGame $(s_t)$
> **end**
>
> *Compute average observed utility for each action*
> $a_t \leftarrow \underset{a}{\arg\max} \; \dfrac{1}{N_a} \displaystyle\sum_{i: a_i = a} r_i$
>
> *Execute selected action in game*
> $s_{t+1} \leftarrow T(s' \,|\, s_t, a_t)$

**end**

---

---

**procedure**  SimulateGame $(s_t)$

---

**for** $u = t \ldots \tau$ **do**

> *Compute Q function approximation*
> $Q^\pi(s_u, a) = \vec{w} \cdot \vec{f}(s_u, a)$
>
> *Sample action from action-value function in $\epsilon$-greedy fashion:*
> $$a_u \sim \pi(s_u, a) = \begin{cases} \text{uniform} \,(a \in A) & \text{with probability } \epsilon \\ \underset{a}{\arg\max} \; Q^\pi(s_u, a) & \text{otherwise} \end{cases}$$
>
> *Execute selected action in game:*
> $s_{u+1} \leftarrow T(s' \,|\, s_u, a_u)$
>
> **if** game is won or lost
> > **break**

**end**

*Update parameters $\vec{w}$ of $Q^\pi(s_t, a)$*

*Return action and observed utility:*
**return** $a_t, R(s_\tau)$

---

Algorithm 1: The general Monte-Carlo algorithm.

### 5.1.1 Modeling Sentence Relevance

As discussed in Section 1, only a small fraction of a strategy document is likely to provide guidance relevant to the current game context. Therefore, to effectively use information from a given document $d$, we first need to identify the sentence $y_i$ that is most relevant to the current game state $s$ and action $a$.[6] We model this decision as a log-linear distribution, defining the probability of $y_i$ being the relevant sentence as:

$$p(y = y_i|s, a, d) \;\propto\; e^{\vec{u} \cdot \vec{\phi}(y_i, s, a, d)}. \tag{1}$$

Here $\vec{\phi}(y_i, s, a, d) \in \mathbb{R}^n$ is a feature function, and $\vec{u}$ are the parameters we need to estimate. The function $\vec{\phi}(\cdot)$ encodes features that combine the attributes of sentence $y_i$ with the attributes of the game state and action. These features allow the model to learn correlations between game attributes and the attributes of relevant sentences.

### 5.1.2 Modeling Predicate Structure

When using text to guide action selection, in addition to using word-level correspondences, we would also like to leverage information encoded in the structure of the sentence. For example, verbs in a sentence might be more likely to describe suggested game actions. We aim to access this information by inducing a task-centric predicate structure on the sentences. That is, we label the words of a sentence as either *action-description*, *state-description* or *background*. Given sentence $y$ and its precomputed dependency parse $q$, we model the word-by-word labeling decision in a log-linear fashion – i.e., the distribution over the predicate labeling $z$ of sentence $y$ is given by:

$$
\begin{aligned}
p(z \,|\, y, q) &= p(\vec{e}\,|\, y, q) \\
&= \prod_j p(e_j | j, y, q), \\
p(e_j | j, y, q) &\propto\; e^{\vec{v} \cdot \vec{\psi}(e_j, j, y, q)},
\end{aligned}
\tag{2}
$$

where $e_j$ is the predicate label of the $j^{\text{th}}$ word. The feature function $\vec{\psi}(e_j, j, y, q) \in \mathbb{R}^n$, in addition to encoding word type and part-of-speech tag, also includes dependency parse information for each word. These features allow the predicate labeling decision to condition on the syntactic structure of the sentence.

### 5.1.3 Modeling the Action-Value Function

Once the relevant sentence has been identified and labeled with a predicate structure, our algorithm needs to use this information along with the attributes of the current game state $s$ to select the best possible game action $a$. To this end, we redefine the action-value function $Q(s, a)$ as a weighted linear combination of features of the game and the text information:

$$Q(s', a') = \vec{w} \cdot \vec{f}(s, a, y_i, z_i). \tag{3}$$

---

6. We use the approximation of selecting the single most relevant sentence as an alternative to combining the features of all sentences in the text, weighted by their relevance probability $p(y = y_i|s, a, d)$. This setup is computationally more expensive than the one used here.

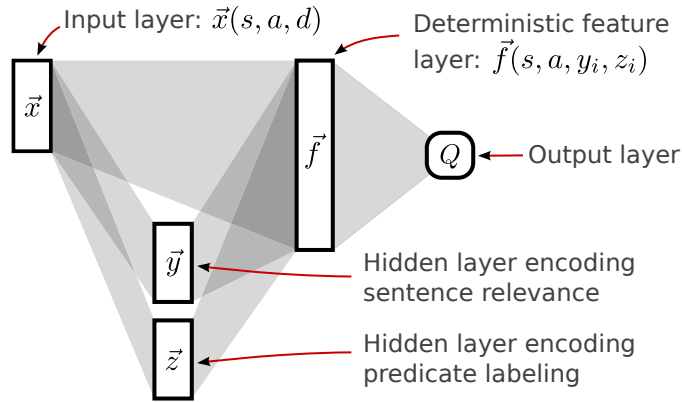

Figure 4: The structure of our neural network model. Each rectangle represents a collection of units in a layer, and the shaded trapezoids show the connections between layers. A fixed, real-valued feature function $\vec{x}(s, a, d)$ transforms the game state $s$, action $a$, and strategy document $d$ into the input vector $\vec{x}$. The second layer contains two disjoint sets of hidden units $\vec{y}$ and $\vec{z}$, where $\vec{y}$ encodes the sentence relevance decisions, and $\vec{z}$ the predicate labeling. These are softmax layers, where only one unit is active at any time. The units of the third layer $\vec{f}(s, a, y_i, z_i)$ are a set of fixed real valued feature functions on $s$, $a$, $d$ and the active units $y_i$ and $z_i$ of $\vec{y}$ and $\vec{z}$ respectively.

Here $s' = \langle s, d \rangle$, $a' = \langle a, y_i, z_i \rangle$, $\vec{w}$ is the weight vector, and $\vec{f}(s, a, y_i, z_i) \in \mathbb{R}^n$ is a feature function over the state $s$, action $a$, relevant sentence $y_i$, and its predicate labeling $z_i$. This structure of the action-value function allows it to explicitly learn the correlations between textual information, and game states and actions. The action $a^*$ that maximizes $Q(s, a)$ is then selected as the best action for state $s$:[7]

$$a^* = \arg\max_a Q(s, a).$$

### 5.1.4 Complete Joint Model

The two text analysis models, and the action-value function described above form the three primary components of our text-aware game playing algorithm. We construct a single principled model from these components by representing each of them via different layers of the multi-layer neural network shown in Figure 4. Essentially, the text analysis decisions are modeled as latent variables by the second, hidden layer of the network, while the final output layer models the action-value function.

---

7. Note that we select action $a^*$ based on $Q(s, a)$, which depends on the relevant sentence $y_i$. This sentence itself is selected conditioned on action $a$. This may look like a cyclic dependency between actions and sentence relevance. However, that is not the case since $Q(s, a)$, and therefore sentence relevance $p(y|s, a, d)$, is computed for every candidate action $a \in A$. The actual game action $a^*$ is then selected from this estimate of $Q(s, a)$.

The *input layer* $\vec{x}$ of our neural network encodes the inputs to the model – i.e., the current state $s$, candidate action $a$, and document $d$. The *second layer* consists of two disjoint sets of hidden units $\vec{y}$ and $\vec{z}$, where each set operates as a stochastic 1-of-$n$ softmax selection layer (Bridle, 1990). The activation function for units in this layer is the standard softmax function:

$$p(y_i = 1 | \vec{x}) = e^{\vec{u}_i \cdot \vec{x}} \Big/ \sum_k e^{\vec{u}_k \cdot \vec{x}},$$

where $y_i$ is the $i^{\text{th}}$ hidden unit of $\vec{y}$, $\vec{u}_i$ is the weight vector corresponding to $y_i$, and $k$ is the number of units in the layer. Given that this activation function is mathematically equivalent to a log-linear distribution, the layers $\vec{y}$ and $\vec{z}$ operate like log-linear models. Node activation in such a softmax layer simulates sampling from the log-linear distribution. We use layer $\vec{y}$ to replicate the log-linear model for sentence relevance from Equation (1), with each node $y_i$ representing a single sentence. Similarly, each unit $z_i$ in layer $\vec{z}$ represents a complete predicate labeling of a sentence, as in Equation (2).[8]

The third *feature layer* $\vec{f}$ of the neural network is deterministically computed given the active units $y_i$ and $z_i$ of the softmax layers, and the values of the input layer. Each unit in this layer corresponds to a fixed feature function $f_k(s, a, y_i, z_i) \in \mathbb{R}$. Finally the *output layer* encodes the action-value function $Q(s, a)$ as a weighted linear combination of the units of the feature layer, thereby replicating Equation (3) and completing the joint model.

As an example of the kind of correlations learned by our model, consider Figure 5. Here, a relevant sentence has already been selected for the given game state. The predicate labeling of this sentence has identified the words "irrigate" and "settler" as describing the action to take. When game roll-outs return higher rewards for the *irrigate* action of the *settler* unit, our model can learn an association between this action and the words that describe it. Similarly, it can learn the association between state description words and the feature values of the current game state – e.g., the word "city" and the binary feature *near-city*. This allows our method to leverage the automatically extracted textual information to improve game play.

## 5.2 Parameter Estimation

Learning in our method is performed in an online fashion: at each game state $s_t$, the algorithm performs a simulated game roll-out, observes the outcome of the simulation, and updates the parameters $\vec{u}$, $\vec{v}$ and $\vec{w}$ of the action-value function $Q(s_t, a_t)$. As shown in Figure 3, these three steps are repeated a fixed number of times at each actual game state. The information from these roll-outs is then used to select the actual game action. The algorithm relearns all the parameters of the action-value function for every new game state $s_t$. This specializes the action-value function to the subgame starting from $s_t$. Learning a specialized $Q(s_t, a_t)$ for each game state is common and useful in games with complex state spaces and dynamics, where learning a single global function approximation can be particularly difficult (Sutton et al., 2007). A consequence of this function specialization is the need for online learning – since we cannot predict which games states will be seen during testing, function specialization for those states cannot be done a priori, ruling out the traditional training/test separation.

---

8. Our intention is to incorporate, into action-value function, information from only the most relevant sentence. Therefore, in practice, we only perform a predicate labeling of the sentence selected by the relevance component of the model.

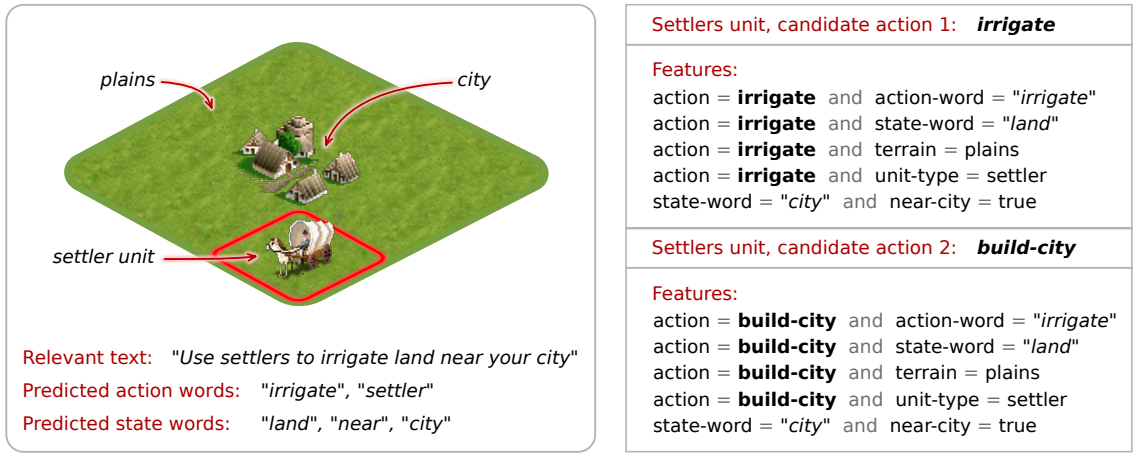

Figure 5: An example of text and game attributes, and resulting candidate action features. On the left is a portion of a game state with arrows indicating game attributes. Also on the left is a sentence relevant to the game state along with action and state words identified by predicate labeling. On the right are two candidate actions for the *settler* unit along with the corresponding features. As mentioned in the relevant sentence, *irrigate* is the better of the two actions – executing it will lead to future higher game scores. This feedback and the features shown above allow our model to learn effective mappings – such as between the action-word "irrigate" and the action *irrigate*, and between state-word "city" and game attribute *near-city*.

Since our model is a non-linear approximation of the underlying action-value function of the game, we learn model parameters by applying non-linear regression to the observed final utilities from the simulated roll-outs. Specifically, we adjust the parameters by stochastic gradient descent, to minimize the mean-squared error between the action-value $Q(s, a)$ and the final utility $R(s_\tau)$ for each observed game state $s$ and action $a$. The resulting update to model parameters $\theta$ is of the form:

$$\Delta \theta = -\frac{\alpha}{2} \nabla_\theta \left[ R(s_\tau) - Q(s, a) \right]^2$$
$$= \alpha \left[ R(s_\tau) - Q(s, a) \right] \nabla_\theta Q(s, a; \theta),$$

where $\alpha$ is a learning rate parameter. This minimization is performed via standard error backpropagation (Bryson & Ho, 1969; Rumelhart, Hinton, & Williams, 1986), resulting in the following online parameter updates:

$$\vec{w} \leftarrow \vec{w} + \alpha_w \left[ Q - R(s_\tau) \right] \vec{f}(s, a, y_i, z_j),$$
$$\vec{u}_i \leftarrow \vec{u}_i + \alpha_u \left[ Q - R(s_\tau) \right] \hat{Q} \vec{x} \left[ 1 - p(y_i | \cdot) \right],$$
$$\vec{v}_i \leftarrow \vec{v}_i + \alpha_v \left[ Q - R(s_\tau) \right] \hat{Q} \vec{x} \left[ 1 - p(z_i | \cdot) \right].$$

Here $\alpha_w$ is the learning rate, $Q = Q(s, a)$, and $\vec{w}$, $\vec{u}_i$ and $\vec{v}_i$ are the parameters of the final layer, the sentence relevance layer and the predicate labeling layer respectively. The derivations of these update equations are given in Appendix A

## 6. Applying the Model

The game we test our model on, Civilization II, is a multi-player strategy game set either on Earth or on a randomly generated world. Each player acts as the ruler of one civilization, and starts with a few game units – i.e., two *Settlers*, two *Workers* and one *Explorer*. The goal is to expand your civilization by developing new technologies, building cities and new units, and to win the game by either controlling the entire world, or successfully sending a spaceship to another world. The map of the game world is divided into a grid of typically 4000 squares, where each grid location represents a tile of either land or sea. Figure 6 shows a portion of this world map from a particular instance of the game, along with the game units of one player. In our experiments, we consider a two-player game of Civilization II on a map of 1000 squares – the smallest map allowed on Freeciv. This map size is used by both novice human players looking for an easier game, as well as advanced players wanting a game of shorter duration. We test our algorithms against the built-in AI player of the game, with the difficulty level at the default *Normal* setting.[9]

### 6.1 Game States and Actions

We define the game state for Monte-Carlo search, to be the map of the game world, along with the attributes of each map tile, and the location and attributes of each player's cities and units. Some examples of these attributes are shown in Figure 7. The space of possible actions for each city and unit is defined by the game rules given the current game state. For example, cities can construct buildings such as harbors and banks, or create new units of various types; while individual units can move around on the grid, and perform unit specific actions such as *irrigation* for *Settlers*, and *military defense* for *Archers*. Since a player controls multiple cities and units, the player's action space at each turn is defined by the combination of all possible actions for those cities and units. In our experiments, on average, a player controls approximately 18 units with each unit having 15 possible actions. The resulting action space for a player is very large – i.e., $10^{21}$. To effectively deal with this large action space, we assume that given the state, the actions of each individual city and unit are independent of the actions of all other cities and units of the same player.[10] At the same time, we maximize parameter sharing by using a single action-value function for all the cities and units of the player.

---

9. Freeciv has five difficulty settings: *Novice*, *Easy*, *Normal*, *Hard* and *Cheating*. As evidenced by discussions on the game's online forum (http://freeciv.wikia.com/index.php?title=Forum:Playing_Freeciv), some human players new to the game find even the *Novice* setting too hard.
10. Since each player executes game actions in turn, i.e. opposing units are fixed during an individual player's turn, the opponent's moves do not enlarge the player's action space.

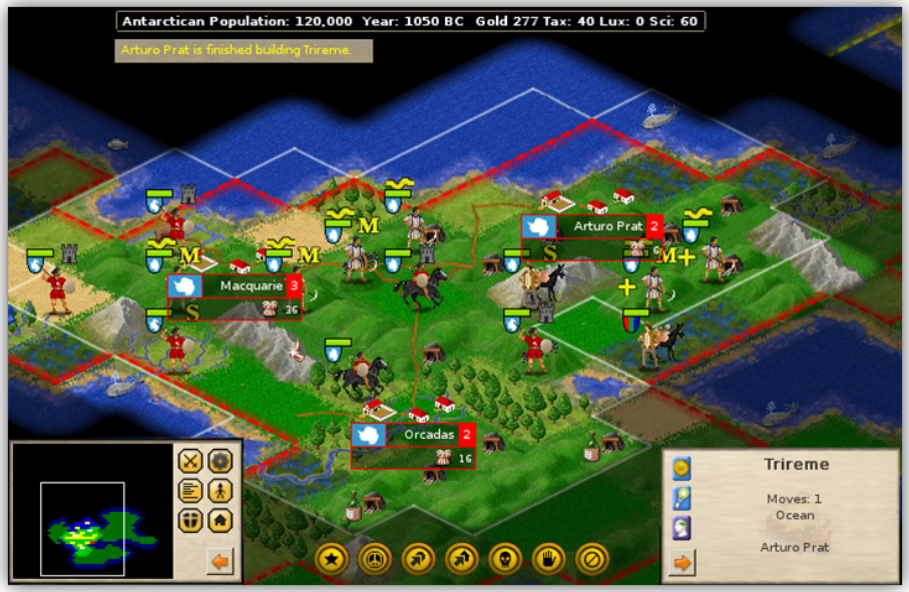

Figure 6: A portion of the game map from one instance of a Civilization II game. Three cities, and several units of a single player are visible on the map. Also visible are the different terrain attributes of map tiles, such as grassland, hills, mountains and deserts.

**Nation attributes:**

- Amount of gold in treasury
- % of world controlled
- Number of cities
- Population
- Known technologies

**Map tile attributes:**

- Terrain type (e.g. grassland, mountain, etc)
- Tile resources (e.g. wheat, coal, wildlife, etc)
- Tile has river
- Construction on tile (city, road, rail, etc)
- Types of units (own or enemy) present

**City attributes:**

- City population
- Surrounding terrain and resources
- Amount of food & resources produced
- Number of units supported by city
- Number & type of units present

**Unit attributes:**

- Unit type (e.g., worker, explorer, archer, etc)
- Unit health & hit points
- Unit experience
- Is unit in a city?
- Is unit fortified?

Figure 7: Example attributes of game state.

## 6.2 Utility Function

Critically important to the Monte-Carlo search algorithm, is the availability of a utility function that can evaluate the outcomes of simulated game roll-outs. In the typical application of the algorithm, the final game outcome in terms of victory or loss is used as the utility function (Tesauro & Galperin, 1996). Unfortunately, the complexity of Civilization II, and the length of a typical game, precludes the possibility of running simulation roll-outs until game completion. The game, however, provides each player with a real valued *game score*, which is a noisy indicator of the strength of their civilization. Since we are playing a two-player game, our player's score relative to the opponent's can be used as the utility function. Specifically, we use the ratio of the game score of the two players.[11]

## 6.3 Features

All the components of our method operate on features computed over a basic set of text and game attributes. The text attributes include the words of each sentence along with their parts-of-speech and dependency parse information such as dependency types and parent words. The basic game attributes encode game information available to human players via the game's graphical user interface. Some examples of these attributes are shown in Figure 7.

To identify the sentence most relevant to the current game state and candidate action, the sentence relevance component computes features over the combined basic attributes of the game and of each sentence from the text. These features $\vec{\phi}$, are of two types – the first computes the Cartesian product between the attributes of the game and the attributes of the candidate sentence. The second type consists of binary features that test for overlap between words from the candidate sentence, and the text labels of the current game state and candidate action. Given that only 3.2% of word tokens from the manual overlap with labels from the game, these similarity features are highly sparse. However, they serve as signposts to guide the learner – as shown by our results, our method is able to operate effectively even in the absence of these features, but performs better when they are present.

Predicate labeling, unlike sentence relevance, is purely a language task and as such operates only over the basic text attributes. The features for this component, $\vec{\psi}$, compute the Cartesian product of the candidate predicate label with the word's type, part-of-speech tag, and dependency parse information. The final component of our model, the action-value approximation, operates over the attributes of the game state, the candidate action, the sentence selected as relevant, and the predicate labeling of that sentence. The features of this layer, $\vec{f}$, compute a three way Cartesian product between the attributes of the candidate action, the attributes of the game state, and the predicate labeled words of the relevant sentence. Overall, $\vec{\phi}$, $\vec{\psi}$ and $\vec{f}$ compute approximately 158,500, 7,900, and 306,800 features respectively – resulting in a total of 473,200 features for our full model. Figure 8 shows some examples of these features.

---

11. The difference between players' scores can also be used as the utility function. However, in practice the score ratio produced better empirical performance across all algorithms and baselines.

Figure 8: Some examples of features used in our model. In each feature, conditions that test game attributes are highlighted in blue, and those that test words in the game manual are highlighted in red.

## 7. Experimental Setup

In this section, we describe the datasets, evaluation metrics, and experimental framework used to test the performance of our method and the various baselines.

### 7.1 Datasets

We use the official game manual of Civilization II as our strategy guide document.[12] The text of this manual uses a vocabulary of 3638 word types, and is composed of 2083 sentences, each on average 16.9 words long. This manual contains information about the rules of the game, about the game user interface, and basic strategy advice about different aspects of the game. We use the Stanford parser (de Marneffe, MacCartney, & Manning, 2006), under default settings, to generate the dependency parse information for sentences in the game manual.

### 7.2 Experimental Framework

To apply our method to the Civilization II game, we use the game's open source reimplementation *Freeciv*.[13] We instrumented FreeCiv to allow our method to programmatically control the game – i.e., to measure the current game state, to execute game actions, to save/load the current game state, and to start and end games.[14]

---

12. www.civfanatics.com/content/civ2/reference/Civ2manual.zip
13. http://freeciv.wikia.com. Game version 2.2

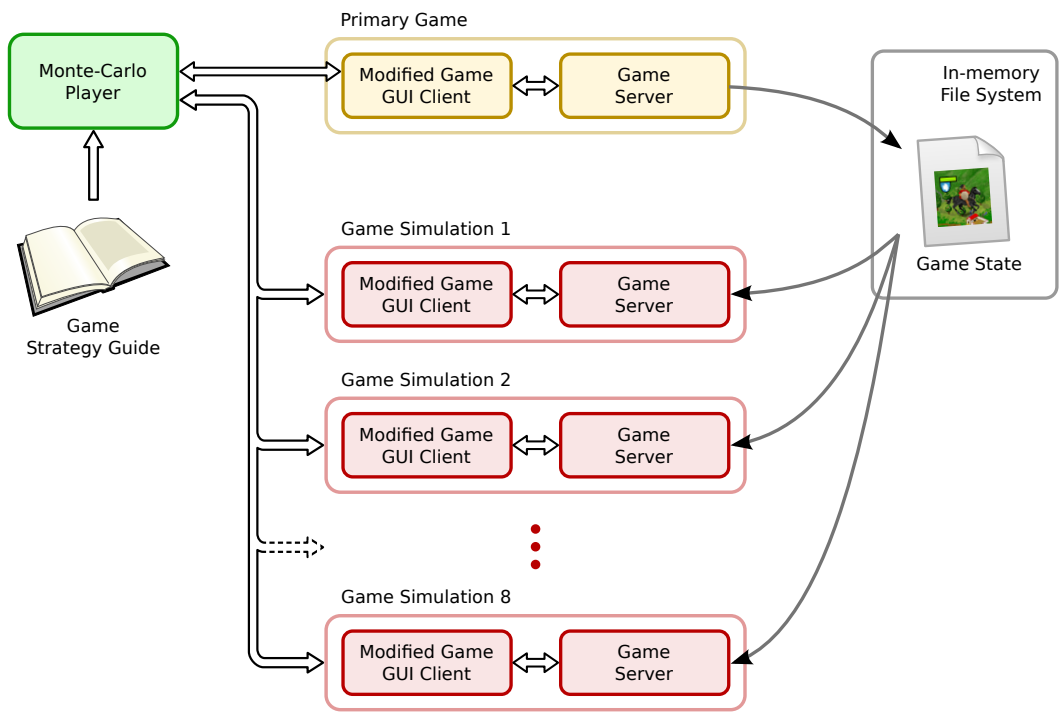

Figure 9: A diagram of the experimental framework, showing the Monte-Carlo player, the server for the primary game which the playing aims to win, and multiple game servers for simulated play. Communications between the multiple processes comprising the framework is via UNIX sockets and an in-memory file system.

Across all experiments, we start the game at the same initial state and run it for 100 steps. At each step, we perform 500 Monte-Carlo roll-outs. Each roll-out is run for 20 simulated game steps before halting the simulation and evaluating the outcome. Note that at each simulated game step, our algorithm needs to select an action for each game unit. Given an average number of units per player of 18, this results in 180,000 decisions during the 500 roll-outs. The pairing of each of these decisions with the corresponding roll-out outcome is used as a datapoint to update model parameters. We use a fixed learning rate of 0.0001 for all experiments. For our method, and for each of the baselines, we run 200 independent games in the above manner, with evaluations averaged across the 200 runs. We use the same experimental settings across all methods, and all model parameters are initialized to zero.

Our experimental setup consists of our Monte-Carlo player, a primary game which we aim to play and win, and a set of simulation games. Both the primary game and the simula-

---

14. In addition to instrumentation, the code of FreeCiv (both the server and client) was changed to increase simulation speed by several orders of magnitude, and to remove bugs which caused the game to crash. To the best of our knowledge, the game rules and functionality are identical to the unmodified Freeciv version 2.2 tions are simply separate instances of the Freeciv game. Each instance of the Freeciv game is made up of one server process, which runs the actual game, and one client process, which is controlled by the Monte-Carlo player. At the start of each roll-out, the simulations are initialized with the current state of the primary game via the game save/reload functionality of Freeciv. Figure 9 shows a diagram of this experimental framework.

The experiments were run on typical desktop PCs with single Intel Core i7 CPUs (4 hyper-threaded cores per CPU). The algorithms were implemented to execute 8 simulation roll-outs in parallel by connecting to 8 independent simulation games. In this computational setup, approximately 5 simulation roll-outs are executed per second for our full model, and a single game of 100 steps runs in 3 hours. Since we treat the Freeciv game code as a black box, special care was taken to ensure consistency across experiments: all code was compiled on one specific machine, under a single fixed build environment (gcc 4.3.2); and all experiments were run under identical settings on a fixed set of machines running a fixed OS configuration (Linux kernel 2.6.35-25, libc 2.12.1).

## 7.3 Evaluation Metrics

We wish to evaluate two aspects of our method: how well it improves game play by leveraging textual information, and how accurately it analyzes text by learning from game feedback. We evaluate the first aspect by comparing our method against various baselines in terms of the percentage of games won against the built-in AI of Freeciv. This AI is a fixed heuristic algorithm designed using extensive knowledge of the game, with the intention of challenging human players.[15] As such, it provides a good open-reference baseline. We evaluate our method by measuring the percentage of games won, averaged over 100 independent runs. However, full games can sometimes last for multiple days, making it difficult to do an extensive analysis of model performance and contributing factors. For this reason, our primary evaluation measures the percentage of games won within the first 100 game steps, averaged over 200 independent runs. This evaluation is an underestimate of model performance – any game where the player has not won by gaining control of the entire game map within 100 steps is considered a loss. Since games can remain tied after 100 steps, two equally matched average players, playing against each other, will most likely have a win rate close to zero under this evaluation.

## 8. Results

To adequately characterize the performance of our method, we evaluate it with respect to several different aspects. In this section, we first describe its game playing performance and analyze the impact of textual information. Then, we investigate the quality of the text analysis produced by our model in terms of both sentence relevance and predicate labeling.

| Method | % Win | % Loss | Std. Err. |
|--------|-------|--------|-----------|
| Random | 0 | 100 | — |
| Built-in AI | 0 | 0 | — |
| Game only | 17.3 | 5.3 | ± 2.7 |
| Latent variable | 26.1 | 3.7 | ± 3.1 |
| **Full model** | **53.7** | 5.9 | ± 3.5 |
| Randomized text | 40.3 | 4.3 | ± 3.4 |

Table 1: Win rate of our method and several baselines within the first 100 game steps, while playing against the built-in game AI. Games that are neither won nor lost are still ongoing. Our model's win rate is statistically significant against all baselines. All results are averaged across 200 independent game runs. The standard errors shown are for percentage wins.

| Method | % Wins | Standard Error |
|--------|--------|----------------|
| Game only | 24.8 | ± 4.3 |
| Latent variable | 31.5 | ± 4.6 |
| **Full model** | **65.4** | ± 4.8 |

Table 2: Win rate of our method and two text-unaware baselines against the built-in AI. All results are averaged across 100 independent game runs.

## 8.1 Game Performance

Table 1 shows the performance of our method and several baselines on the primary 100-step evaluation. In this scenario, our language-aware Monte-Carlo algorithm wins on average 53.7% of games, substantially outperforming all baselines, while the best non-language-aware method has a win rate of only 26.1%. The dismal performance of the *Random* baseline and the game's own *Built-in AI*, playing against itself, are indications of the difficulty of winning games within the first 100 steps. As shown in Table 2, when evaluated on full length games, our method has a win rate of 65.4% compared to 31.5% for the best text-unaware baseline.[16]

---

15. While this AI is constrained to follow the rules of the game, it has access to information typically not available to human players, such as information about the technology, cities and units of it's opponents. Our methods on the other hand are restricted to the actions and information available to human players.

16. Note that the performance of all methods on the full games is different from those listed in our previous publications (Branavan, Silver, & Barzilay, 2011a, 2011b). These previous numbers were biased by a code flaw in FreeCiv which caused the game to sporadically crash in the middle game play. While we originally believed the crash to be random, it was subsequently discovered to happen more often in losing games, and thereby biasing the win rates of all methods upwards. The numbers presented here are with this game bug fixed, with no crashes observed in any of the experiments.

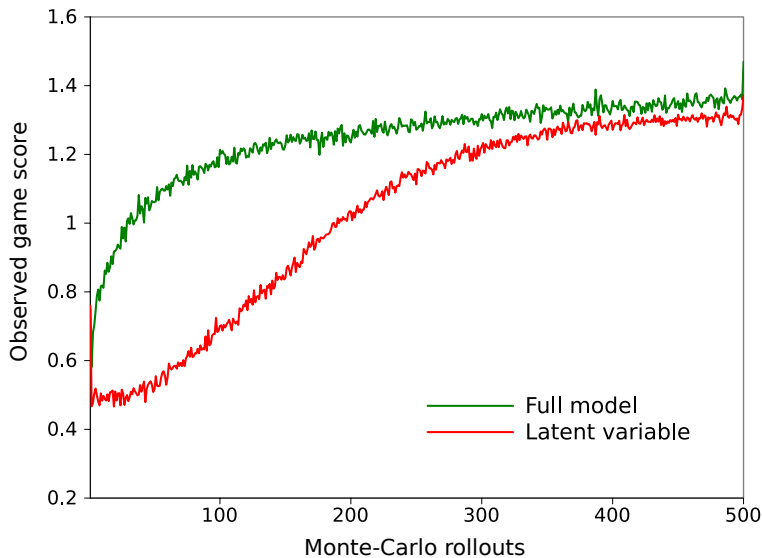

Figure 10: Observed game score as a function of Monte-Carlo roll-outs for our text-aware full model, and the text-unaware latent-variable model. Model parameters are updated after each roll-out, thus performance improves with roll-outs. As can be seen, our full model's performance improves dramatically over a small number of roll-outs, demonstrating the benefit it derives from textual information.

### 8.1.1 Textual Advice and Game Performance

To verify and characterize the impact of textual advice on our model's performance, we compare it against several baselines that do not have access to textual information. The simplest of these methods, *Game only*, models the action-value function $Q(s, a)$ as a linear approximation of the game's state and action attributes. This non-text-aware method wins only 17.3% of games (see Table 1). To confirm that our method's improved performance is not simply due to its inherently richer non-linear approximation, we also evaluate two ablative non-linear baselines. The first of these, *Latent variable* extends the linear action-value function of *Game only* with a set of latent variables. It is in essence a four layer neural network, similar to our full model, where the second layer's units are activated only based on game information. This baseline wins 26.1% of games (Table 1), significantly improving over the linear *Game only* baseline, but still trailing our text-aware method by more than 27%. The second ablative baseline, *Randomized text*, is identical to our model, except that it is given a randomly generated document as input. We generate this document by randomly permuting the locations of words in the game manual, thereby maintaining the document's statistical properties in terms of type frequencies. This ensures that the number of latent variables in this baseline is equal to that of our full model. Thus, this baseline has a model capacity equal to our text-aware method while not having access to any textual information. The performance of this baseline, which wins only 40.3% of games, confirms that information extracted from text is indeed instrumental to the performance of our method.

Figure 10 provides insight into how textual information helps improve game performance – it shows the observed game score during the Monte-Carlo roll-outs for our full model and the latent-variable baseline. As can be seen from this figure, the textual information guides our model to a high-score region of the search space far quicker than the non-text aware method, thus resulting in better overall performance. To evaluate how the performance of our method varies with the amount of available textual-information, we conduct an experiment where only random portions of the text are given to the algorithm. As shown in Figure 11, our method's performance varies linearly as a function of the amount of text, with the *Randomized text* experiment corresponding to the point where no information is available from text.

### 8.1.2 Impact of Seed Vocabulary on Performance

The sentence relevance component of our model uses features that compute the similarity between words in a sentence, and the text labels of the game state and action. This assumes the availability of a seed vocabulary that names game attributes. In our domain, of the 256 unique text labels present in the game, 135 occur in the vocabulary of the game manual. This results in a sparse seed vocabulary of 135 words, covering only 3.7% of word types and 3.2% of word tokens in the manual. Despite this sparsity, the seed vocabulary can have a potentially large impact on model performance since it provides an initial set of word groundings. To evaluate the importance of this initial grounding, we test our method with an empty seed vocabulary. In this setup, our full model wins 49.0% of games, showing that while the seed words are important, our method can also operate effectively in their absence.

### 8.1.3 Linguistic Representation and Game Performance

To characterize the contribution of language to game performance, we conduct a series of evaluations which vary the type and complexity of the linguistic analysis performed by our method. The results of this evaluation are shown in Table 3. The first of these, *Sentence relevance*, highlights the contributions of the two language components of our model. This algorithm, which is identical to our full model but lacks the predicate labeling component, wins 46.7% of games, showing that while it is essential to identify the textual advice relevant to the current game state, a deeper syntactic analysis of the extracted text substantially improves performance.

To evaluate the importance of dependency parse information in our language analysis, we vary the type of features available to the predicate labeling component of our model. The first of these ablative experiments, *No dependency information*, removes all dependency features – leaving predicate labeling to operate only on word type features. The performance of this baseline, a win rate of 39.6%, clearly shows that the dependency features are crucial for model performance. The remaining three methods – *No dependency label*, *No dependency parent POS tag* and *No dependency parent word* – each drop the dependency feature they are named after. The contribution of these features to model performance can be seen in Table 3.

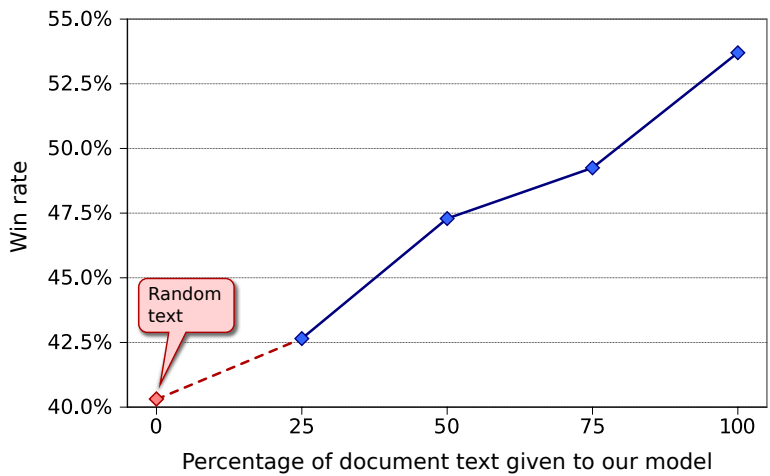

Figure 11: The performance of our text-aware model as a function of the amount of text available to it. We construct partial documents by randomly sub-sampling sentences from the full game manual. The *x*-axis shows the amount of sentences given to the method as a ratio of the full text. At the leftmost extreme is the performance of the *Randomized Text* baseline, showing how it fits into the performance trend at the point of having no useful textual information.

| Method | % Win | % Loss | Std. Err. |
|---|---|---|---|
| **Full model** | **53.7** | 5.9 | ± 3.5 |
| Sentence relevance | 46.7 | 2.8 | ± 3.5 |
| No dependency information | 39.6 | 3.0 | ± 3.4 |
| No dependency label | 50.1 | 3.0 | ± 3.5 |
| No depend. parent POS tag | 42.6 | 4.0 | ± 3.5 |
| No depend. parent word | 33.0 | 4.0 | ± 3.3 |

Table 3: Win rates of several ablated versions of our model, showing the contribution of different aspects of textual information to game performance. *Sentence relevance* is identical to the *Full model*, except that it lacks the predicate labeling component. The four methods at the bottom of the table ablate specific dependency features (as indicated by the method's name) from the predicate labeling component of the full model.

### 8.1.4 Model Complexity vs Computation Time Trade-off

One inherent disadvantage of non-linear models, when compared to simpler linear models, is the increase in computation time required for parameter estimation. In our Monte-Carlo Search setup, model parameters are re-estimated after each simulated roll-out. Therefore, given a fixed amount of time, more roll-outs can be done for a simpler and faster model. By its very nature, the performance of Monte-Carlo Search improves with the number of roll-outs. This trade-off between model complexity and roll-outs is important since a simpler model could compensate by using more roll-outs, and thereby outperform more complex ones. This scenario is particularly relevant in games where players have a limited amount of time for each turn.

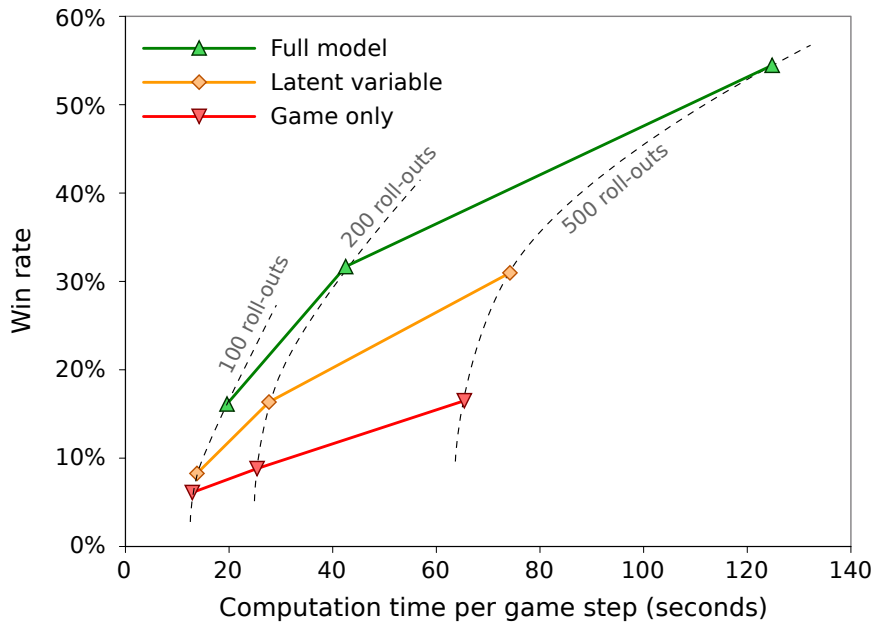

Figure 12: Win rate as a function of computation time per game step. For each Monte-Carlo search method, win rate and computation time were measured for 100, 200 and 500 roll-outs per game step, respectively.

To explore this trade-off, we vary the number of simulation roll-outs allowed for each method at each game step, recording the win-rate and the average computation time per game. Figure 12 shows the results of this evaluation for 100, 200 and 500 roll-outs. While the more complex methods have higher computational demands, these results clearly show that even when given a fixed amount of computation time per game step, our text-aware model still produces the best performance by a wide margin.

### 8.1.5 LEARNED GAME STRATEGY

Qualitatively, all of the methods described here learn a basic *rush strategy*. Essentially, they attempt to develop basic technologies, build an army, and take over opposing cities as quickly as possible. The performance difference between the different models is essentially due to how well they learn this strategy.

There are two basic reasons why our algorithms learn the rush strategy. First, since we are attempting to maximize game score, the methods are implicitly biased towards finding the fastest way to win – which happens to be the rush strategy when playing against the built-in AI of Civilization 2. Second, more complex strategies typically require the coordination of multiple game units. Since our models assume game units to be independent, they cannot explicitly learn such coordination – putting many complex strategies beyond the capabilities of our algorithms.

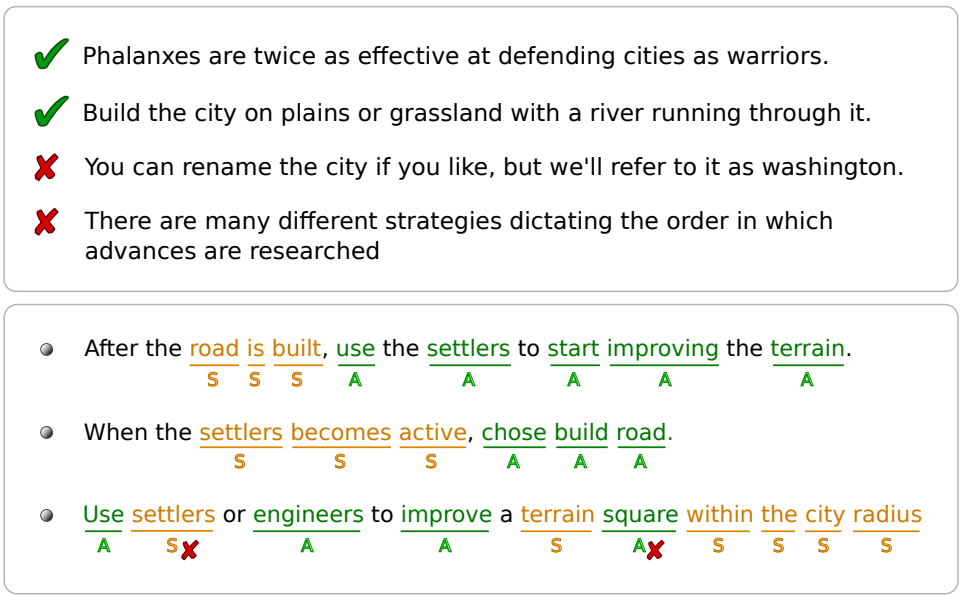

Figure 13: Examples of our method's sentence relevance and predicate labeling decisions. The box above shows two sentences (identified by green check marks) which were predicted as relevant, and two which were not. The box below shows the predicted predicate structure of three sentences, with "S" indicating *state* description, "A" *action* description and *background* words unmarked. Mistakes are identified with crosses.

## 8.2 Accuracy of Linguistic Analysis

As described in Section 5, text analysis in our method is tightly coupled with game playing – both in terms of modeling, and in terms of learning from game feedback. We have seen from the results thus far, that this text analysis does indeed help game play. In this section we focus on the game-driven text analysis itself, and investigate how well it conforms to more common notions of linguistic correctness. We do this by comparing model predictions of sentence relevance and predicate labeling against manual annotations.

### 8.2.1 Sentence Relevance

Figure 13 shows examples of the sentence relevance decisions produced by our method. To evaluate the accuracy of these decisions, we would ideally like to use a ground-truth relevance annotation of the game's user manual. This however, is impractical since the relevance decision is dependent on the game context, and is hence specific to each time step of each game instance. Therefore, we evaluate sentence relevance accuracy using a synthetic document. We create this document by combining the original game manual with an equal number of sentences which are known to be irrelevant to the game. These sentences are collected by randomly sampling from the Wall Street Journal corpus (Marcus, Santorini, & Marcinkiewicz, 1993).[17] We evaluate sentence relevance on this synthetic document by measuring the accuracy with which game manual sentences are picked as relevant.

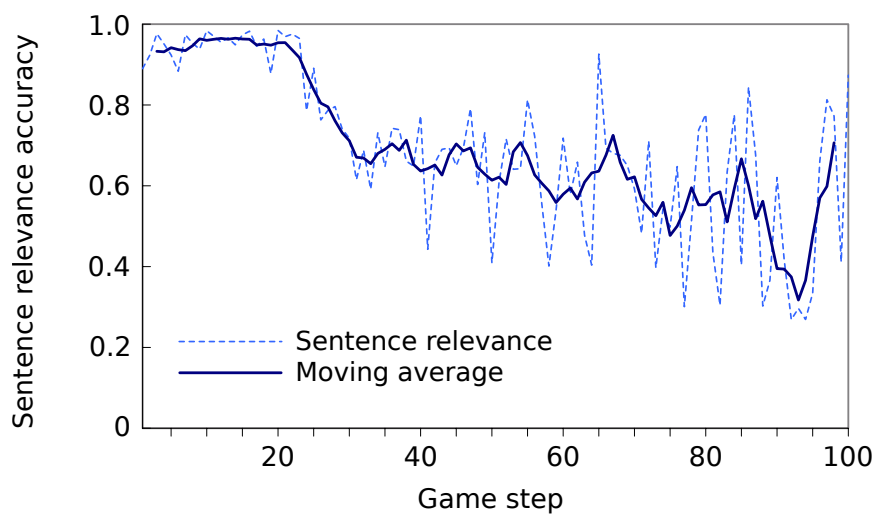

Figure 14: Accuracy of our method's sentence relevance predictions, averaged over 100 independent runs.

In this evaluation, our method achieves an average accuracy of 71.8%. Given that our model only has to differentiate between the game manual text and the Wall Street Journal, this number may seem disappointing. Furthermore, as can be seen from Figure 14, the sentence relevance accuracy varies widely as the game progresses, with a high average of 94.2% during the initial 25 game steps. In reality, this pattern of high initial accuracy followed by a lower average is not entirely surprising: the official game manual for Civilization II is written for first time players. As such, it focuses on the initial portion of the game, providing little strategy advice relevant to subsequent game play.[18] If this is the reason for the observed sentence relevance trend, we would also expect the final layer of the neural network to emphasize game features over text features after the first 25 steps of the game. This is indeed the case, as can be seen in Figure 15.

To further test this hypothesis, we perform an experiment where the first $n$ steps of the game are played using our full model, and the subsequent $100 - n$ steps are played without using any textual information. The results of this evaluation for several values of $n$ are given in Figure 16, showing that the initial phase of the game is indeed where information from the game manual is most useful. In fact, this hybrid method performs just as well as our full model when $n = 50$, achieving a 53.3% win rate. This shows that our method is able to accurately identify relevant sentences when the information they contain is most pertinent to game play, and most likely to produce better game performance.

---

17. Note that sentences from the WSJ corpus contain words such as *city* which can potentially confuse our algorithm, causing it to select such sentences are relevant to game play.

18. This is reminiscent of *opening books* for games like Chess or Go, which aim to guide the player to a playable middle game, without providing much information about subsequent game play.

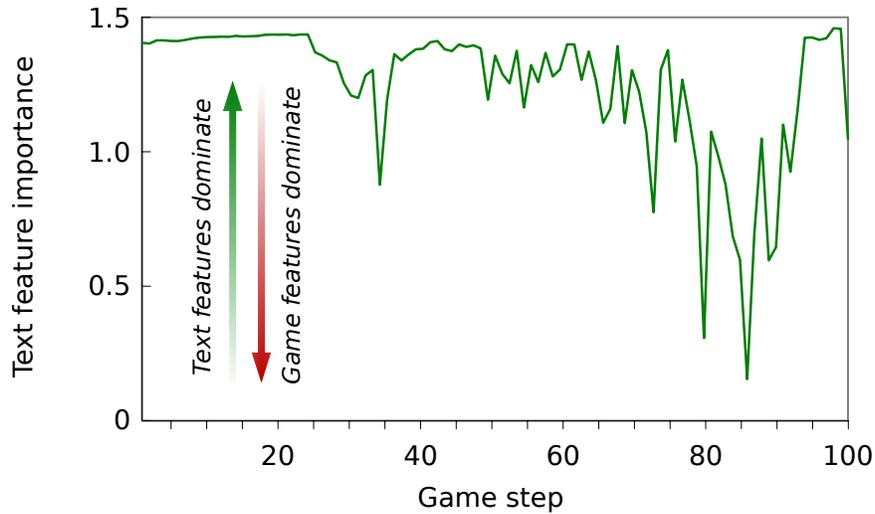

Figure 15: Difference between the norms of the text features and game features of the output layer of the neural network. Beyond the initial 25 steps of the game, our method relies increasingly on game features.

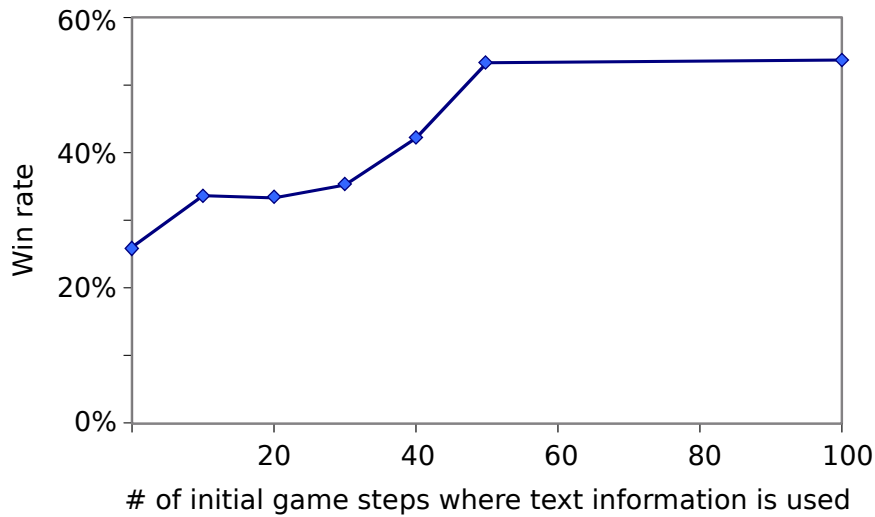

Figure 16: Graph showing how the availability of textual information during the initial steps of the game affects the performance of our full model. Textual information is given to the model for the first *n* steps (the *x* axis), beyond which point the algorithm has no access to text, and becomes equivalent to the *Latent Variable* model – i.e., the best non-text model.

| Method | S/A/B | S/A |
|--------|-------|-----|
| Random labeling | 33.3% | 50.0% |
| Model, first 100 steps | 45.1% | 78.9% |
| Model, first 25 steps | 48.0% | 92.7% |

Table 4: Predicate labeling accuracy of our method and a random baseline. Column "S/A/B" shows performance on the three-way labeling of words as *state*, *action* or *background*, while column "S/A" shows accuracy on the task of differentiating between state and action words.

| game attribute | word |
|----------------|------|
| state: grassland | "city" |
| state: grassland | "build" |
| state: hills | "build" |
| action: settlers_build_city | "city" |
| action: set_research | "discovery" |
| action: settlers_build_city | "settler" |
| action: settlers_goto_location | "build" |
| action: city_build_barracks | "construct" |
| action: research_alphabet | "develop" |
| action: set_research | "discovery" |

Figure 17: Examples of word to game attribute associations that are learned via the feature weights of our model.

### 8.2.2 Predicate Labeling

Figure 13 shows examples of the predicate structure output of our model. We evaluate the accuracy of this labeling by comparing it against a gold-standard annotation of the game manual.[19] Table 4 shows the performance of our method in terms of how accurately it labels words as *state*, *action* or *background*, and also how accurately it differentiates between *state* and *action* words. In addition to showing a performance improvement over the random baseline, these results display a clear trend: under both evaluations, labeling accuracy is higher during the initial stages of the game. This is to be expected since the model relies heavily on textual features during the beginning of the game (see Figure 15).

To verify the usefulness of our method's predicate labeling, we perform a final set of experiments where predicate labels are selected uniformly at random within our full model. This random labeling results in a win rate of 44% – a performance similar to the *sentence relevance* model which uses no predicate information. This confirms that our method is able to identify a predicate structure which, while noisy, provides information relevant to game play. Figure 17 shows examples of how this textual information is grounded in the game, by way of the associations learned between words and game attributes in the final layer of the full model. For example, our model learns a strong association between the game-state attribute *grassland* and the words "city" and "build", indicating that textual information about building cities maybe very useful when a player's unit is near grassland.

---

19. Note that a ground truth labeling of words as either *action-description*, *state-description*, or *background* is based purely on the semantics of the sentence, and is independent of game state. For this reason, manual annotation is feasible, unlike in the case of sentence relevance.

## 9. Conclusions

In this paper we presented a novel approach for improving the performance of control applications by leveraging information automatically extracted from text documents, while at the same time learning language analysis based on control feedback. The model biases the learned strategy by enriching the policy function with text features, thereby modeling the mapping between words in a manual and state-specific action selection. To effectively learn this grounding, the model identifies text relevant to the current game state, and induces a predicate structure on that text. These linguistic decisions are modeled jointly using a non-linear policy function trained in the Monte-Carlo Search framework.

Empirical results show that our model is able to significantly improve game win rate by leveraging textual information when compared to strong language-agnostic baselines. We also demonstrate that despite the increased complexity of our model, the knowledge it acquires enables it to sustain good performance even when the number of simulations is reduced. Moreover, deeper linguistic analysis, in the form of a predicate labeling of text, further improves game play. We show that information about the syntactic structure of text is crucial for such an analysis, and ignoring this information has a large impact on model performance. Finally, our experiments demonstrate that by tightly coupling control and linguistic features, the model is able to deliver robust performance in the presence of the noise inherent in automatic language analysis.

### Bibliographical Note

Portions of this work were previously presented in two conference publications (Branavan et al., 2011a, 2011b). This article significantly extends our previous work, most notably by providing an analysis of model properties such as the impact of linguistic representation on model performance, dependence of the model on bootstrapping conditions, and the trade-off between the model's representational power and its empirical complexity (Section 8). The paper also significantly increases the volume of experiments on which we base our conclusions. In addition, we provide a comprehensive description of the model, providing full mathematical derivations supporting the algorithm (Section 5.1 and Appendix A).

### Acknowledgments

The authors acknowledge the support of the NSF (CAREER grant IIS-0448168, grant IIS-0835652), the DARPA BOLT Program (HR0011-11-2-0008), the DARPA Machine Reading Program (FA8750-09-C-0172, PO#4910018860), Batelle (PO#300662) and the Microsoft Research New Faculty Fellowship. Thanks to the anonymous reviewers, Michael Collins, Tommi Jaakkola, Leslie Kaelbling, Nate Kushman, Sasha Rush, Luke Zettlemoyer, and the MIT NLP group for their suggestions and comments. Any opinions, findings, conclusions, or recommendations expressed in this paper are those of the authors, and do not necessarily reflect the views of the funding organizations.

## Appendix A. Parameter Estimation

The parameter of our model are estimated via standard error backpropagation (Bryson & Ho, 1969; Rumelhart et al., 1986). To derive the parameter updates, consider the slightly simplified neural network shown below. This network is identical to our model, but for the sake of clarity, it has only a single second layer $\vec{y}$ instead of the two parallel second layers $\vec{y}$ and $\vec{z}$. The parameter updates for these parallel layers $\vec{y}$ and $\vec{z}$ are similar, therefore we will show the derivation only for $\vec{y}$ in addition to the updates for the final layer.

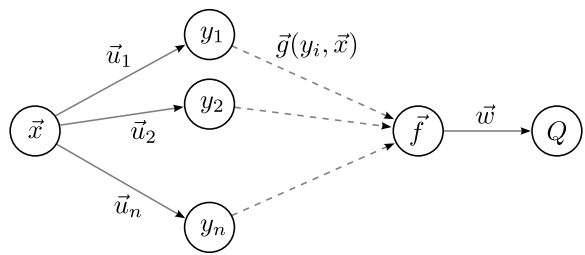

As in our model, the nodes $y_i$ in the network above are activated via a softmax function; the third layer, $\vec{f}$, is computed deterministically from the active nodes of the second layer via the function $\vec{g}(y_i, \vec{x})$; and the output $Q$ is a linear combination of $\vec{f}$ weighted by $\vec{w}$:

$$
\begin{aligned}
p(y_i = 1 \mid \vec{x}; \vec{u}_i) &= \frac{e^{\vec{u}_i \cdot \vec{x}}}{\sum_k e^{\vec{u}_k \cdot \vec{x}}}, \\
\vec{f} &= \sum_i \vec{g}(\vec{x}, y_i) \, p(y_i \mid \vec{x}; \vec{u}_i), \\
Q &= \vec{w} \cdot \vec{f}.
\end{aligned}
$$

Our goal is to minimize the mean-squared error $e$ by gradient descent. We achieve this by updating model parameters along the gradient of $e$ with respect to each parameter. Using $\theta_i$ as a general term to indicate our model's parameters, this update takes the form:

$$
\begin{aligned}
e &= \frac{1}{2}(Q - R)^2, \\
\Delta \theta_i &= \frac{\partial e}{\partial \theta_i} \\
&= (Q - R) \, \frac{\partial Q}{\partial \theta_i}.
\end{aligned} \tag{4}
$$

From Equation (4), the updates for final layer parameters are given by:

$$
\begin{aligned}
\Delta w_i &= (Q - R) \, \frac{\partial Q}{\partial w_i} \\
&= (Q - R) \, \frac{\partial}{\partial w_i} \vec{w} \cdot \vec{f} \\
&= (Q - R) \, f_i.
\end{aligned}
$$

Since our model samples the one most relevant sentence $y_i$, and the best predicate labeling $z_i$, the resulting online updates for the output layer parameters $\vec{w}$ are:

$$\vec{w} \;\leftarrow\; \vec{w} + \alpha_w \; [Q - R(s_\tau)] \; \vec{f}(s, a, y_i, z_j),$$

where $\alpha_w$ is the learning rate, and $Q = Q(s, a)$. The updates for the second layer's parameters are similar, but somewhat more involved. Again, from Equation (4),

$$
\begin{aligned}
\Delta u_{i,j} &= (Q - R) \; \frac{\partial Q}{\partial u_{i,j}} \\
&= (Q - R) \; \frac{\partial}{\partial u_{i,j}} \; \vec{w} \cdot \vec{f} \\
&= (Q - R) \; \frac{\partial}{\partial u_{i,j}} \; \vec{w} \cdot \sum_k \vec{g}(\vec{x}, y_k) \; p(y_i \mid \vec{x}; \vec{u}_k) \\
&= (Q - R) \; \vec{w} \cdot \vec{g}(\vec{x}, y_i) \; \frac{\partial}{\partial u_{i,j}} \; p(y_i \mid \vec{x}; \vec{u}_i).
\end{aligned}
\tag{5}
$$

Considering the final term in the above equation separately,

$$
\begin{aligned}
\frac{\partial}{\partial u_{i,j}} \; p(y_i \mid \vec{x}; \vec{u}_i) &= \frac{\partial}{\partial u_{i,j}} \; \frac{e^{\vec{u}_i \cdot \vec{x}}}{Z}, \qquad \text{where } Z = \sum_k e^{\vec{u}_k \cdot \vec{x}} \\
&= \left(\frac{e^{\vec{u}_i \cdot \vec{x}}}{Z}\right) \frac{\frac{\partial}{\partial u_{i,j}} \frac{e^{\vec{u}_i \cdot \vec{x}}}{Z}}{\left(\frac{e^{\vec{u}_i \cdot \vec{x}}}{Z}\right)} \\
&= \left(\frac{e^{\vec{u}_i \cdot \vec{x}}}{Z}\right) \frac{\partial}{\partial u_{i,j}} \log\left[\frac{e^{\vec{u}_i \cdot \vec{x}}}{Z}\right] \\
&= \left(\frac{e^{\vec{u}_i \cdot \vec{x}}}{Z}\right) \left[x_j - \frac{\partial}{\partial u_{i,j}} \log Z\right] \\
&= \left(\frac{e^{\vec{u}_i \cdot \vec{x}}}{Z}\right) \left[x_j - \frac{1}{Z} \frac{\partial Z}{\partial u_{i,j}}\right] \\
&= \left(\frac{e^{\vec{u}_i \cdot \vec{x}}}{Z}\right) \left[x_j - \frac{1}{Z} \frac{\partial}{\partial u_{i,j}} \sum_k e^{\vec{u}_k \cdot \vec{x}}\right] \\
&= \left(\frac{e^{\vec{u}_i \cdot \vec{x}}}{Z}\right) \left[x_j - \frac{1}{Z} x_j e^{\vec{u}_k \cdot \vec{x}}\right] \\
&= \left(\frac{e^{\vec{u}_i \cdot \vec{x}}}{Z}\right) x_j \left[1 - \frac{e^{\vec{u}_i \cdot \vec{x}}}{Z}\right].
\end{aligned}
$$

Therefore, from Equation (5),

$$
\begin{aligned}
\Delta u_{i,j} &= (Q-R) \ \vec{w} \cdot \vec{g}(\vec{x}, y_i) \ \frac{\partial}{\partial u_{i,j}} \ p(y_i \mid \vec{x}; \vec{u}_i) \\
&= (Q-R) \ \vec{w} \cdot \vec{g}(\vec{x}, y_i) \ \left( \frac{e^{\vec{u}_i \cdot \vec{x}}}{Z} \right) x_j \ \left[ 1 - \frac{e^{\vec{u}_i \cdot \vec{x}}}{Z} \right] \\
&= (Q-R) \ x_j \ \vec{w} \cdot \vec{g}(\vec{x}, y_i) \ p(y_i \mid \vec{x}; \vec{u}_i) \ [1 - p(y_i \mid \vec{x}; \vec{u}_i)] \\
&= (Q-R) \ x_j \ \hat{Q} \ [1 - p(y_i \mid \vec{x}; \vec{u}_i)] \,,
\end{aligned}
$$

where $\hat{Q} = \vec{w} \cdot \vec{g}(\vec{x}, y_i) \ p(y_i \mid \vec{x}; \vec{u}_i)$.

The resulting online updates for the sentence relevance and predicate labeling parameters $\vec{u}$ and $\vec{v}$ are:

$$
\begin{aligned}
\vec{u}_i &\leftarrow \vec{u}_i + \alpha_u \ [Q - R(s_\tau)] \ \hat{Q} \ \vec{x} \ [1 - p(y_i | \cdot)], \\
\vec{v}_i &\leftarrow \vec{v}_i + \alpha_v \ [Q - R(s_\tau)] \ \hat{Q} \ \vec{x} \ [1 - p(z_i | \cdot)].
\end{aligned}
$$

## Appendix B. Example of Sentence Relevance Predictions

Shown below is a portion of the strategy guide for Civilization II. Sentences that were identified as relevant by our text-aware model are highlighted in green.

*Choosing your location.*

When building a new city, carefully plan where you place it. ==Citizens can work the terrain surrounding the city square in an x-shaped pattern (see city radius for a diagram showing the exact dimensions).== This area is called the city radius (the terrain square on which the settlers were standing becomes the city square). The natural resources available where a population settles affect its ability to produce food and goods. ==Cities built on or near water sources can irrigate to increase their crop yields, and cities near mineral outcroppings can mine for raw materials. On the other hand, cities surrounded by desert are always handicapped by the aridness of their terrain, and cities encircled by mountains find arable cropland at a premium.== In addition to the economic potential within the city's radius, you need to consider the proximity of other cities and the strategic value of a location. Ideally, you want to locate cities in areas that offer a combination of benefits : food for population growth, raw materials for production, and river or coastal areas for trade. Where possible, take advantage of the presence of special resources on terrain squares (see terrain & movement for details on their benefits).

*Strategic value.*

The strategic value of a city site is a final consideration. ==A city square's underlying terrain can increase any defender's strength when that city comes under attack.== In some circumstances, the defensive value of a particular city's terrain might be more important than the economic value; consider the case where a continent narrows to a bottleneck and a rival holds the other side. ==Good defensive terrain (hills, mountains, and jungle) is generally poor for food production and inhibits the early growth of a city. If you need to compromise between growth and defense, build the city on a plains or grassland square with a river running through it if possible.== This yields decent trade production and gains a 50 percent defense bonus. ==Regardless of where a city is built, the city square is easier to defend than the same unimproved terrain. In a city you can build the city walls improvement, which triples the defense factors of military units stationed there.== Also, units defending a city square are destroyed one at a time if they lose. Outside of cities, all units stacked together are destroyed when any military unit in the stack is defeated (units in fortresses are the only exception; see fortresses). ==Placing some cities on the seacoast gives you access to the ocean. You can launch ship units to explore the world and to transport your units overseas. With few coastal cities, your sea power is inhibited.==

## Appendix C. Examples of Predicate Labeling Predictions

Listed below are the predicate labellings computed by our text-aware method on example sentences from the game manual. The predicted labels are indicated below the words with the letters A, S, and B for *action-description*, *state-description* and *background* respectively. Incorrect labels are indicated by a red check mark, along with the correct label in brackets.

- After the road is built, use the settlers to start improving the terrain.
  S  S  S  A  A  A  A  A

- When the settlers becomes active, chose build road.
  S  S  S  A  A  A

- Use settlers or engineers to improve a terrain square within the city radius
  A  ✗S(A)  A  A  S  ✗A(S)  S  S  S  S

- Bronze working allows you to build phalanx units
  S  S  ✗B(S)  A  A  B

- In order to expand your civilization , you need to build other cities
  S  ✗A(S)  S  B  A  ✗B(A)

- In order to protect the city , the phalanx must remain inside
  ✗B(S)  S  ✗B(S)  A  ✗S(A)  A  ✗B(A)

- As soon as you've found a decent site , you want your settlers to build a
  ✗B(S)  S  ✗B(S)  S  ✗A(B)  ✗B(A)  A

  permanent settlement - a city
  ✗S(A)  A

- In a city you can build the city walls improvement
  ✗A(S)  ✗B(A)  A  A  A

- Once the city is undefended , you can move a friendly army into the city and capture it
  S  S  ✗B(S)  A  A  A  A  B

- You can build a city on any terrain square except for ocean.
  A  ✗S(A)  ✗B(S)  S  ✗A(S)  S

- You can launch ship units to explore the world and to transport your units overseas
  A  ✗S(A)  S  ✗B(S)  ✗B(S)  S  B

- When a city is in disorder, disband distant military units, return them to their home cities,
  ✗A(S)  A  ✗S(A)  A  A  ✗S(A)

  or change their home cities
  A  A  A

- You can build a wonder only if you have discovered the advance that makes it possible
  A  ✗S(A)  S  S  S  S

## Appendix D. Examples of Learned Text to Game Attribute Mappings

Shown below are examples of some of the word to game-attribute associations learned by our model. The top ten game attributes with the strongest association by feature weight are listed for three of the example words – "attack", "build" and "grassland". For the fourth word, "settler", only seven attributes had non-zero weights in experiments used to collect these statistics.

**attack**

phalanx  (unit)
warriors  (unit)
colossus  (wonder)
city walls  (city improvement)
archers  (unit)
catapult  (unit)
palace  (city improvement)
coinage  (city production)
city_build_warriors  (action)
city_build_phalanx  (action)

**build**

worker_goto  (action)
settler_autosettle  (action)
worker_autosettle  (action)
pheasant  (terrain attribute)
settler_irrigate  (action)
worker_mine  (action)
build_city_walls  (action)
build_catapult  (action)
swamp  (terrain attribute)
grassland  (terrain attribute)

**grassland**

settler_build_city  (action)
worker_continue_action  (action)
pheasant  (terrain attribute)
city_build_improvement  (action)
city_max_production  (action)
settlers  (state attribute)
city_max_food  (action)
settler_goto  (action)
worker_build_road  (action)
pyramids  (city attribute)

**settler**

settlers  (state attribute)
settler_build_city  (action)
city  (state_attribute)
grassland  (terrain_attribute)
plains  (terrain_attribute)
road  (terrain_attribute)
workers  (state attribute)

## Appendix E. Features Used by the Model

**Features used predict sentence relevance**

The following templates are used to compute the features for sentence relevance:

- *Word W is present in sentence.*
- *Number of words that match the text label of the current unit, an attribute in the immediate neighborhood of the unit, or the action under consideration.*
- *The unit's type is U, (e.g., `worker`) and word W is present in sentence.*
- *The action type is A, (e.g., `irrigate`) and word W is present in sentence.*

**Features used predict predicate structure**

The following templates are used to compute the features for the predicate labeling of words. The label being considered for the word (i.e., `action`, `state` or `background`) is denoted by L.

- *Label is L and the word type is W.*
- *Label is L and the part-of-speech tag of the word is T.*
- *Label is L and the parent word in the dependency tree is W.*
- *Label is L and the dependency type to the dependency parent word is D.*
- *Label is L and the part-of-speech of the dependency parent word is T.*
- *Label is L and the word is a leaf node in the dependency tree.*
- *Label is L and the word is not a leaf node in the dependency tree.*
- *Label is L and the word matches a state attribute name.*
- *Label is L and the word matches a unit type name.*
- *Label is L and the word matches a action name.*

**Features used to model action-value function**

The following templates are used to compute the features of the action-value approximation. Unless otherwise mentioned, the features look at the attributes of the player controlled by our model.

- *Percentage of world controlled.*
- *Percentage of world explored.*
- *Player's game score.*
- *Opponent's game score.*
- *Number of cities.*
- *Average size of cities.*
- *Total size of cities.*

- *Number of units.*
- *Number of veteran units.*
- *Wealth in gold.*
- *Excess food produced.*
- *Excess shield produced.*
- *Excess trade produced.*
- *Excess science produced.*
- *Excess gold produced.*
- *Excess luxury produced.*
- *Name of technology currently being researched.*
- *Percentage completion of current research.*
- *Percentage remaining of current research.*
- *Number of game turns before current research is completed.*

The following feature templates are applied to each city controlled by the player:

- *Current size of city.*
- *Number of turns before city grows in size.*
- *Amount of food stored in city.*
- *Amount of shield stored in city ("shields" are used to construct new buildings and units in the city).*
- *Turns remaining before current construction is completed.*
- *Surplus food production in city.*
- *Surplus shield production in city.*
- *Surplus trade production in city.*
- *Surplus science production in city.*
- *Surplus gold production in city.*
- *Surplus luxury production in city.*
- *Distance to closest friendly city.*
- *Average distance to friendly cities.*
- *City governance type.*
- *Type of building or unit currently under construction.*
- *Types of buildings already constructed in city.*
- *Type of terrain surrounding the city.*
- *Type of resources available in the city's neighborhood.*

- *Is there another city in the neighborhood.*
- *Is there an enemy unit in the neighborhood.*
- *Is there an enemy city in the neighborhood.*

The following feature templates are applied to each unit controlled by the player:

- *Type of unit.*
- *Moves left for unit in current game turn.*
- *Current health of unit.*
- *Hit-points of unit.*
- *Is unit a veteran.*
- *Distance to closest friendly city.*
- *Average distance to friendly cities.*
- *Type of terrain surrounding the unit.*
- *Type of resources available in the unit's neighborhood.*
- *Is there an enemy unit in the neighborhood.*
- *Is there an enemy city in the neighborhood.*

The following feature templates are applied to each predicate-labeled word in the sentence selected as relevant, combined with the current state and action attributes:

- *Word W is present in sentence, and the action being considered is A.*
- *Word W with predicate label P is present in sentence, and the action being considered is A.*
- *Word W is present in sentence, the current unit's type is U, and the action being considered is A.*
- *Word W with predicate label P is present in sentence, the current unit's type is U, and the action being considered is A.*
- *Word W is present in sentence, and the current unit's type is U.*
- *Word W with predicate label P is present in sentence, and the current unit's type is U.*
- *Word W is present in sentence, and an attribute with text label A is present in the current unit's neighborhood.*
- *Word W with predicate label P is present in sentence, and an attribute with text label A is present in the current unit's neighborhood.*